# Proceedings of the 1ˢᵗ Workshop on Robotics Challenges and Vision (RCV2013)

June 27, 2013

## Table of Contents

# Program Committee

| | |
|---|---|
| Nancy Amato | Texas A & M University |
| Hamidreza Chitsaz | Wayne State University |
| Howie Choset | Carnegie Mellon University |
| Kamal Gupta | Simon Fraser University |
| Seth Hutchinson | University of Illinois at Urbana-Champaign |
| Lydia Kavraki | Rice University |
| Moslem Kazemi | Carnegie Mellon University |
| Dinesh Manocha | University of North Carolina-Chapel Hill |

## Best Paper Awards

| | |
|---|---|
| 1st | Soft Subdivision Search in Motion Planning |
| 2nd | Large Motion Libraries: Toward a "Google" for Robot Motions |
| 3rd | Predicting the Change - A Step Towards Life-Long Operation in Everyday Environments |

# Soft Subdivision Search in Motion Planning

Chee K. Yap

Courant Institute, New York University
New York, NY 10012 USA
Email: yap@cs.nyu.edu

*Abstract*—The main paradigm for practical motion planning in the last two decades is Probabilistic Road Map (PRM). We propose an alternative paradigm called *Soft Subdivision Search* (SSS). The SSS approach is based on two ingredients: the standard subdivision of space, coupled with *soft predicates*. Such predicates are conservative and convergent relative to exact predicates. This leads to a new class of *resolution-exact* planners.

We view PRM and SSS as frameworks for broad classes of planners. There are many parallels between SSS and PRM: both frameworks are versatile, practical, easy to implement, with adaptive local complexity. The critical difference is that SSS avoids the Halting Problem of PRM. We address three issues:
(1) We axiomatize some basic properties that allow resolution-exact planners to be constructed in the SSS framework.
(2) We show how soft predicates can be effectively and correctly implemented using numerical approximations.
(3) We recover exact planners by extending our framework.

The SSS framework is a theoretically sound basis for new classes of algorithms in motion planning and beyond. We discuss the prospects of SSS planners being able to solve currently challenging problems and their relation to PRM.

## I. Introduction

Motion Planning is a fundamental problem in robotics. It originated as the "findpath problem" in Artificial Intelligence [3]. In the 1980s, computational geometers began the algorithmic study of motion planning [19, 9], focusing on *exact planners*: such planners return a path if any exists, and report "No Path" otherwise. It was early observed that there are two universal approaches to motion planning: Cell Decomposition [16] and Retraction [19]. After the work of Canny [4], the retraction is popularly known as the "roadmap approach". In the 1990's the roadmap approach takes another turn.

**¶1. Theory.** Today, exact motion planning is textbook material [6, 13] and continues to be actively investigated (e.g., [8]). Nevertheless its impact on robotics is modest: Zhang et al. [22] noted that exact implementations have been limited to 3 degrees of freedom, and for simple robots only. Exactness is costly as it implicitly requires algebraic numbers. The techniques of Exact Geometric Computation [20] can avoid direct manipulation of algebraic numbers, and is practical for many basic problems. Nevertheless, the usual expedient is to replace exact arithmetic by approximate machine arithmetic, leading to the ubiquitous problems of numerical non-robustness. Efficiency aside, there is a fundamental but less well-known barrier: *the Turing computability of exact algorithms for most non-algebraic problems is unknown* [5]. This barrier exists in most problems beyond kinematic motion planning. See [5] for a rare non-algebraic planning problem that is computable using transcendental number theory.

**¶2. Practice.** Since the mid 1990's, the Probabilistic Road Map (PRM) paradigm has been dominant; Kavraki et al. [11] gave the basic formulation. A more general viewpoint [12] regards PRM as the probabilistic form of "Sampling Road Maps". For simplicity, we use "PRM" as surrogate for all sampling methods. Quoting Choset et al [6, p.201]: "*PRM, EST, RRT, SRT, and their variants have changed the way path planning is performed for high-dimensional robots. They have also paved the way for the development of planners for problems beyond basic path planning.*" In his invited talk at the workshop[1] on open problems in this field, J.C. Latombe stated that the major open problem of PRM is that it does not know how to terminate when there is no path. In practice, the algorithm is timed-out, but this leads to problems such as the "Climbers Dilemma" giving rise to issues like the "Climbers Dilemma" (Bretl 2005). We will call it the **Halting Problem for PRM** (like the Halting Problem in Turing machines, it is a semi-decidable). It is the ultimate form of the "Narrow Passage Problem" [6, p. 201]. Latombe's talk suggested promising approaches such as Lazy PRM [2], but clearly a large part of the literature is devoted to this issue. The theoretical basis for PRM algorithms is probabilistic completeness [10], or more generally, "sampling completeness." But the Halting Problem is inherent in such completeness.

**¶3. Common Ground.** We seek a common ground for theory and practice: stronger theoretical guarantees without the inordinate demands of exactness. Fortunately, exactness is a mismatch for robotics. This is evident from the remark that all physical constants, devices and sensors have limited accuracy. Yet it does not absolve us from mathematical precision if we wish the theoretical development of robotic algorithms to thrive. This tension between practice and theory has led to their divergent paths described above. We turn to the idea of "resolution complete" methods, noting that the early work of Brooks and Lozano-Perez [3] was already on this track. It is known that resolution complete methods can avoid the Halting Problem [22]. The notion of resolution completeness is seldom scrutinized. Our companion paper [17] pointed out some untenable or lacking interpretations, and proposed

---

[1] IROS 2011 Workshop on Progress and Open Problems in Motion Planning, September 30, 2011, San Francisco.

the notion of **resolution-exactness**. Surprisingly, resolution-exactness has "inherent" indeterminacy, even for deterministic algorithms using exact predicates. This indeterminacy is mild compared to sampling completeness, and a good match for the needs of robotics.

Resolution-exactness is basically a numeric/analytic concept, but exactness is not an adjective we associate with numerics. Our main contribution is to explicate ideas that are intuitively known to practitioners, to provide a clear foundation for theoretical algorithm designers to ply their craft. This seems critical: without a clear foundation, many theoreticians would shun such algorithms. Exact computation has served as *the* foundation for over 2 decades, but its limitations are showing (as discussed above). What we need is a viable *exact numerical foundation* (cf. Smale's effort in this direction). We will argue that algorithms in our framework are not just theoretically-sound but implementable and useful.

**¶4. Overview of Paper.** Our full paper [21] aims to expose the foundations of resolution exactness. There are four themes:
(0) We take a leaf from the success of PRM research: the simplicity and generality of PRM framework ensures that implementers of this framework can get easy access to a whole family of algorithms. This led us to formulate an analogous framework called **soft subdivision search** (SSS). Curious aside: PRM and SSS correspond (resp.) roughly to the two universal approaches to motion planning, namely retraction/roadmap and cell-decomposition.
(1) Next, we axiomatize the setting for SSS planners by considering the problem of finding paths in $Y \subseteq X$ connecting given $\alpha, \beta \in Y$ where $X$ is a normed linear space. The boxes in subdivision trees may be shapes such as simplices. The goal is to derive general principles to guide the design of SSS planners. Here we must avoid the temptation of excessive generality, leading to weak generic results about metric spaces. We aim at a balance which captures a large class of problems about which non-trivial theorems can be proved.
(2) One bane of exact algorithms is the "implementation gap". Exact primitives are typically algebraic but implementers use machine arithmetic approximation, thereby forfeiting all the guarantees of exactness. We derive principles for correct implementation of soft predicates, and derive error estimates to allow machine arithmetic filters.
(3) SSS planners take an input resolution parameter $\varepsilon$ which must be positive. If we admit $\varepsilon = 0$, the planners become non-halting like PRM. We indicate solutions, leading to new classes of exact algorithms.

Within the constraints of this workshop presentation, we only address theme (0) and give a critical evaluation of SSS planners for solving challenging planning problems.

## II. BASICS

To aid further discussion, we need some definitions. Our terminology is quite standard, but we rely on readers' intuition for now. Consider the standard kinematic motion planning problem for a fixed rigid robot $R_0 \subseteq \mathbb{R}^k$ ($k = 2, 3$) which defines a configuration space $C_{space} = C_{space}(R_0)$. The planner input is $(\varepsilon, \Omega, \alpha, \beta)$ where $\varepsilon > 0$ is the **resolution parameter**, $\Omega \subseteq \mathbb{R}^k$ is a polyhedral set, and $\alpha, \beta \in C_{space}$. The planner is **resolution-exact** (or $\varepsilon$-**exact**) if it has a constant $K > 1$ such that on any input, it outputs an $\Omega$-avoiding path from $\alpha$ to $\beta$ if there exist paths with clearance $K\varepsilon$; and it outputs "No Path" if there are no paths with clearance $\varepsilon/K$. The role of $K$ is critical: it causes indeterminacy, but is also the key to avoiding exact computation.

There are two ingredients in resolution-based methods: first is the subdivision of $C_{space}$. The subdivision may be organized as a **subdivision tree**, often called quadtrees. Tree nodes correspond to subsets $B \subseteq C_{space}$ with simple shapes such as boxes or simplices. Although grid search is often identified with resolution complete algorithms, we stress that grid methods are usually a form of sampling and inadequate for $\varepsilon$-exactness. The second ingredient is a **classification predicate** to decide if a node $B$ is free or not. The obstacle set $\Omega$ defines the free space, $C_{free} = C_{free}(R_0, \Omega) \subseteq C_{space}$. Wlog, $C_{free}$ is an open set; its boundary $\partial C_{free}$ comprises the **semi-free** configurations. Say $\alpha \in C_{space}$ is **stuck** if it is neither free nor semi-free. The **exact classification predicate** is $C : B \mapsto C(B) \in \{\text{FREE}, \text{STUCK}, \text{MIXED}\}$ such that

$$C(B) = \begin{cases} \text{FREE} & \text{if } B \subseteq C_{free}, \\ \text{STUCK} & \text{if } B \cap \overline{C_{free}} \text{ is empty,} \\ \text{MIXED} & \text{else.} \end{cases}$$

It is easy to construct exact planners using $C$. Moreover $C(B)$ *could* be computed exactly for nice $B$ (e.g., $B$ is a box). But our thrust is to avoid the high cost of exactness. The basic idea (ansatz) is: *in the presence of subdivision, exact predicates can be replaced by suitable approximations*. More precisely, a predicate $\widetilde{C} : B \mapsto \widetilde{C}(B) \in \{\text{FREE}, \text{STUCK}, \text{MIXED}\}$ is a **soft version** of $C$ if it is **conservative** (i.e., $\widetilde{C}(B) \neq \text{MIXED}$ implies $\widetilde{C}(B) = C(B)$), and **convergent** (i.e., $B_i \to p \in C_{space}$ as $i \to \infty$ implies $\widetilde{C}(B_i) \to C(p)$). For analysis, we may also need some measure of "effectivity" or convergence rate.

What we know: soft predicates are relatively easy to design and to implement. Indeed, the computation of $\widetilde{C}$ can be correlated to the expansion of the subdivision tree $\mathcal{T}$. LaValle insightfully call this aspect of our work as "opening up the blackbox of collision testing". Soft predicates have nice properties like **composability**: if a polyhedral robot $R_0 \subseteq \mathbb{R}^k$ has a cover $R_0 = \cup_{i=1}^m T_i$ then we obtain a soft predicate for $R_i$ from soft predicates for the $T_i$'s. Thus predicates for complex robots like $R_0$ is reducible to simpler robots $T_i$. To ensure efficiency and adaptivity, the technique of filters will prove essential (a well-known phenomenon in numerical methods). We refer to related work (not just in motion planning) exemplifying these remarks (e.g., [14, 17, 18, 15]).

**¶5. Critical Discussion.** It is hard to claim novelty in an old idea like resolution-based methods, although we claim new theoretical foundations. It is harder to improve upon a 20-year old paradigm like PRM, where remarkable advances have been

made over the years. Nevertheless, we claim a place for SSS planners in practical robotics under PRM's shadow.

First, we address a conceptual objection. Some critiques view the "No Path" outcome in SSS planners as equivalent to the "PRM time-out" in resolution space. Hence it is no less "arbitrary". This analogy can mislead in two ways. First, the said inherent inaccuracies in sensors, actuators and physical constants mean that paths with clearance below some (calculable) resolution are as good as "No Path". So the "No Path" outcome in SSS planners may be principled, not arbitrary. Second, "time-out" in resolution search is only the most obvious way terminate (we use it below), but it is not the only way. In fact, it is a deeply interesting question to develop techniques for fast detection of "No Path" (cf. [22], [7]).

Consider our suggestion that SSS is practical. The subdivision infrastructure is well-understood and based on efficient data structures like union-find. The soft predicates we design [17] can mostly reduce to estimating distances between two obstacle features (i.e., point, line or plane). This almost seems trivial compared to exact algorithms; so SSS planners are clearly *implementable*. But will these implementations be *practically efficient*? Here, we invoke the evidence of prior resolution-based work such as Zhu and Latombe [23], Barhehenn and Hutchinson [1], and Zhang, Kim and Manocha [22]. Of course, we must reinterpret them using our new perspective; it is illuminating to revisit these papers with hindsight.

The preceding paragraph is not new except for our SSS perspective. The deeper debate among roboticists is about the ability of resolution methods to scale up in dimension. The consensus is that resolution methods can only reach medium degrees-of-freedom (DOF), while PRM reaches much higher[2] degrees. Choset [6, p. 202] suggests that state-of-art PRM can handle DOF in the range $5 - 12$. They noted that a 10 DOF planar robot from Kavraki (1995) cannot be tackled by other methods. However, we find no conceptual barriers for SSS to match PRM. Randomness is not pertinent since SSS can also expand randomly. A naive approach to resolution methods yields tree sizes that are exponential in the depth; but related subdivision work in root isolation [15] shows we can achieve tree size that is worst-case polynomial in the depth. The performance of SSS planners are highly dependent on the tree expansion strategy; this is no different from PRM. Indeed, the tree structure of subdivision seems to give SSS a great advantage. Currently, we have limited (but encouraging) experiments to support our intuition, but we feel the field is wide open for experimentation.

## III. TWO FRAMEWORKS FOR MOTION PLANNING.

We intend to view PRM as an **algorithmic framework** for a large class of sampling-based planners. An algorithm within the framework is[3] just a specific instantiation, using particular data structures and subroutines.

**¶6. The PRM Framework.** There are many possible formulations, but we follow LaValle [13, Section 5.4.1]: to find a path connecting $\alpha, \beta \in C_{free}$, we maintain a graph $G = (V, E)$ where $\{\alpha, \beta\} \subseteq V \subseteq C_{free}$ and edges in $E$ correspond to paths. We need three predicates: $Free(u)$ to test if configuration $u$ is free; $Connect(v, u)$ to test if the (straight) motion from $v$ to $u$ is free; a **termination criterion** that returns "success" (a path is found) or "failure" (time-out or other condition).

---

PRM FRAMEWORK:
While (termination criterion fails):
    1. Vertex Selection Method (VSM):
        Choose a vertex $v$ in $V$ for expansion.
    2. Configuration Generation Method (CGM):
        Generate some $u \in C_{space}$ (perhaps near $v$)
    3. Local Planning Method (LPM):
        If $Free(u)$,
            Add $u$ to $V$
            If $Connect(v, u)$, add $(v, u)$ to $E$.
Return success or failure accordingly.

---

Choset et al. [6, p.198] noted that PRM is practical because the predicate $Free(u)$ is relatively cheap. The large literature on collision detection is about this predicate. We offer another reason for the great success of PRM: *the framework allows one to easily modify the constituent components (VSM, CGM, LPM) to obtain a variety of algorithms for diverse needs. The basic infrastructure is relatively stable, thanks to the simplicity and generality of PRM.* We want to emulate this in SSS.

**¶7. The SSS Framework.** In addition to the usual input $(\varepsilon, \Omega, \alpha, \beta)$, an initial box $B_0 \subseteq C_{space}$ is given: we are interested in paths restricted to $B_0$. The subdivision tree $\mathcal{T}$ is rooted at $B_0$, and each node $B \subseteq B_0$ is classified by a soft predicate $\widetilde{C}$. We grow $\mathcal{T}$ by expanding successive MIXED-leaves until we find a path or conclude "No Path".

---

SSS FRAMEWORK
1. While ($\widetilde{C}(Box(\alpha)) \neq$ FREE)    ◁ *Initialization*
    If $Box(\alpha)$ has length $< \varepsilon$, Return ("No Path")
    Else **Expand**($Box(\alpha)$)
   While ($\widetilde{C}(Box(\beta)) \neq$ FREE)
    ... do the same for $\beta$ ...
2. While ($Find(Box(\alpha)) \neq Find(Box(\beta))$)    ◁ *Main Loop*
    If $Q$ is empty, Return("No Path")
    $B \leftarrow Q.$GetNext()
    **Expand**($B$)
3. Compute a FREE channel $P$ from $Box(\alpha)$ to $Box(\beta)$
    Generate and return the "canonical path" $\overline{P}$ inside $P$.

---

A priority queue $Q = Q_{\mathcal{T}}$ holds all MIXED-leaves of radius $r(B) \geq \varepsilon$. The routine $Q.$**GetNext**() returns a leaf of highest priority which is split by **Expand**(B). The FREE boxes are stored in a union-find structure for the connected components: two boxes $B, B'$ are directly connected if $\dim(B \cap B') = d-1$. If "$Box(\alpha)$" is a leaf of $\mathcal{T}$ containing $\alpha$, our algorithm halts as soon as $Box(\alpha)$ and $Box(\beta)$ are in the same component.

The performance of SSS is naturally adaptive but highly dependent on GetNext() and Expand(). For $d \geq 4$ degrees of

---

freedom, careful expansion is critical; do not always expect to split into $2^d$ children. The methods in [23, 1, 22] fall under our framework.

**¶8. Similarities and Differences.** There are many similarities between PRM and SSS, especially in their contrasts to exact algorithms. Both have two key subroutines: (i) search strategies (VSM in PRM, GetNext() in SSS), and (ii) freeness testing ($Free(u)$ in PRM, $\widetilde{C}(B)$ in SSS). Both SSS and PRM have the ability to find paths *before* exploring the entire $C_{free}$. Thus, Hsu et al. [10, p. 640] calls this a "foundational choice in PRM planning". In contrast, exact methods require expensive (non-adaptive) pre-processing to compute a description of $C_{free}$. Both frameworks naturally compute a path, i.e., a parametrized curve in $C_{free}$; exact methods require a separate subalgorithm for this.

The key difference is that SSS planners have no Halting Problem. In SSS we use a more demanding predicate $\widetilde{C}(B)$ than $Free(u)$. The benefit is that SSS can compute free channels just by checking adjacency of two FREE boxes; PRM needs an extra predicate $Connect(v, u)$.

## IV. CONCLUSION

In this paper, we described the SSS framework for $\varepsilon$-exact planners. Ideas of resolution-limited algorithms are well-known, but to our knowledge, the simple[4] properties of soft classifiers have never been isolated. These "simple ideas" promise to create new algorithms that are practical *and* theoretically sound, not only in motion planning. There are many open questions concerning SSS. Although there are interesting theoretical questions, we feel the immediate challenge is to prove the practical power of SSS. Following up on [17], we plan to do this by designing and implementing a variety of soft predicates and search strategies, from 6DOF robots, medium DOF flexible robots (e.g., Kavraki robot), and complex robots. Like PRM, we expect many variants of SSS to arise.

### REFERENCES

[1] M. Barbehenn and S. Hutchinson. Toward an exact incremental geometric robot motion planner. In *Proc. Intell. Robots and Systems 95*, vol. 3, pp. 39–44, 1995.

[2] R. Bohlin and L.E. Kavraki. A randomized algorithm for robot path planning based on lazy evaluation. *Handbook on Randomized Comput.*, pp. 221–249. Kluwer, 2001.

[3] R. A. Brooks and T. Lozano-Perez. A subdivision algorithm in configuration space for findpath with rotation. In *8th IJCAI - Vol. 2*, pp. 799–806, San Francisco, 1983.

[4] John Francis Canny. *The complexity of robot motion planning*. The MIT Press, 1988. PhD thesis, M.I.T.

[5] E.-C. Chang, S. W. Choi, D. Kwon, H. Park, and C. Yap. Shortest paths for disc obstacles is computable. *IJCGA*, 16(5-6):567–590, 2006. Special Issue.

[6] H. Choset, K. M. Lynch, S. Hutchinson, G. Kantor, W. Burgard, L. E. Kavraki, and S. Thrun. *Principles of Robot Motion: Theory, Algorithms, and Implementations*. MIT Press, 2005.

[7] J. Denny and N. M. Amato. Toggle PRM: A coordinated mapping of C-free and C-obstacle in arbitrary dimension. In *Proc. WAFR*. MIT, Cambridge, USA. June 2012.

[8] M. Safey el Din and E. Schost. A baby steps/giant steps probabilistic algorithm for computing roadmaps in smooth bounded real hypersurface. *Discrete and Comp. Geom.*, 45(1):181–220, 2011.

[9] D. Halperin, L. Kavraki, and J.-C. Latombe. Robotics. In J. E. Goodman and J. O'Rourke, eds., *Handbook of Discrete and Comp. Geom.*, chap. 41. CRC Press, 1997.

[10] D. Hsu, J.-C. Latombe, and H. Kurniawati. On the probabilistic foundations of probabilistic roadmap planning. *IJRR*, 25(7):627–643, 2006.

[11] L. Kavraki, P. Švestka, C. Latombe, and M. Overmars. Probabilistic roadmaps for path planning in high-dimensional configuration spaces. *IEEE Trans. Robotics and Automation*, 12(4):566–580, 1996.

[12] S. LaValle, M. Branicky, and S. Lindemann. On the relationship between classical grid search and probabilistic roadmaps. *IJRR*, 23(7/8):673–692, 2004.

[13] Steven M. LaValle. *Planning Algorithms*. Cambridge U. Press, 2006.

[14] S. Plantinga and G. Vegter. Isotopic approximation of implicit curves and surfaces. In *Proc. SGP*, pp. 245–254, New York, 2004. ACM Press.

[15] M. Sagraloff and C.K. Yap. A simple but exact and efficient algorithm for complex root isolation. *36th ISSAC*, pp. 353–360, 2011. June 8-11, San Jose, California.

[16] J. T. Schwartz and M. Sharir. On the piano movers' problem: II. General techniques for computing topological properties of real algebraic manifolds. *Advances in Appl. Math.*, 4:298–351, 1983.

[17] C. Wang, Y.-J. Chiang, and C. Yap. On Soft Predicates in Subdivision Motion Planning. In *29th Symp. on Comp. Geom.*. To Appear. Rio de Janeiro. Jun 17-20, 2013.

[18] Chee Yap, Vikram Sharma, and Jyh-Ming Lien. Towards Exact Numerical Voronoi diagrams. In *9th ISVD*, pp. 2–16. IEEE, 2012. Invited Talk.

[19] Chee K. Yap. Algorithmic motion planning. In *Advances in Robotics, Vol. 1: Algorithmic and geometric issues*, vol. 1, pp. 95–143. Lawrence Erlbaum Associates, 1987.

[20] Chee K. Yap. Robust geometric computation. In J. E. Goodman and J. O'Rourke, eds., *Handbook of Discrete and Computational Geometry*, chap. 41, pp. 927–952. Chapman & Hall/CRC, Boca Raton, FL, 2nd ed., 2004.

[21] Chee K. Yap. Theory of Soft Subdivision Search and Motion Planning, 2012. Manuscript. URL http://cs.nyu.edu/exact/.

[22] L. Zhang, Y. J. Kim, and D. Manocha. Efficient cell labelling and path non-existence computation using C-obstacle query. *IJRR*, 27(11–12), 2008.

[23] D.J. Zhu and J.-C. Latombe. New heuristic algorithms for efficient hierarchical path planning. *IEEE Transactions on Robotics and Automation*, 7:9–20, 1991.

---

[4] Some reviewers of our work see only the safeness part of soft classifiers. They fail to note that previous work are silent about convergence or effectivity.

# Large Motion Libraries: Toward a "Google" for Robot Motions

Kris Hauser, School of Informatics and Computing, Indiana University at Bloomington
hauserk@indiana.edu

*Abstract*— There is a growing need in robotics for real-time optimal planning and control, driven by the advent new technologies like autonomous vehicles, legged robot locomotion, object manipulation, CAD/CAM, computer animation, and surgical robots. But even at the current state-of-the-art, global optimization is generally too computationally expensive for real-time use. The status quo appears unsuitable looking ahead to the future, which will require addressing progressively higher dimensional systems, faster response rates, longer time horizons, large and detailed environments, and problems with uncertainty. I propose that a motion library approach has the potential to address these upcoming needs. The idea is to first precompute a large library of motion primitives on a set of representative training environments. The robot will then retrieve primitives online to solve novel problems. Given enough training data and perfect recall, performance is limited only by the retrieval cost. The major challenge to address is scale: how many primitives are needed to generalize across all environments and tasks of interest, and how can tools for precomputation and retrieval scale up to thousands or millions of primitives? In this paper, I present a preliminary roadmap for motion library research that will help move toward a "Google" for robot motions.

*Keywords: robotics; optimization; motion planning; machine learning; information retrieval*

## I. Introduction

For decades robotics has had to cope with the fact that global optimization is painfully slow, even though although local optimization and solving for feasible suboptimal solutions are generally fast. Moore's law can no longer be relied upon to deliver better performance; although memory cost and storage density continues to improve, CPU speeds and energy costs are starting to plateau (arguably serial performance has already plateaued). The implications are pervasive. Whether the object being optimized is a path, a trajectory, a feedback control policy, a grasp, a geometric quantity, etc., a great deal of human effort must be invested to engineer the environment or devise good heuristics (e.g., initial guesses for local optimizers) to calculate high-quality behaviors. As a result, developing intelligent behavior is time-consuming, even in controlled lab settings.

Can robots automatically learn motion strategies and when to use them? The question of "when to use a motion" is a major challenge, because it requires mapping the space of *problems* (i.e., initial conditions, tasks, and environments) to the space of *optimal motions*. In the worst case, this mapping is intractably complex, but it may be the case that the map can be tractably approximated. For example, the empirical distribution of problems might be approximated by a finite sample, and that problem features are statistically highly correlated with optimized motions. If the world obeys such a structure, the following motion library approach may be useful (Figure 1):

1. **Precompute** a huge number of problem/optimal motion pairs (motion primitives) and store them in a database (a motion library). Problems will be generated from a set of representative training environments, or from perceptual inputs gathered offline.

2. Online, the robot solves novel problems by **retrieving** appropriate primitives from the library according to a problem similarity metric, and **adapting** them. (More complex forms of adaptation might compose multiple primitives together in sequence through high-level planning, or by blending.)

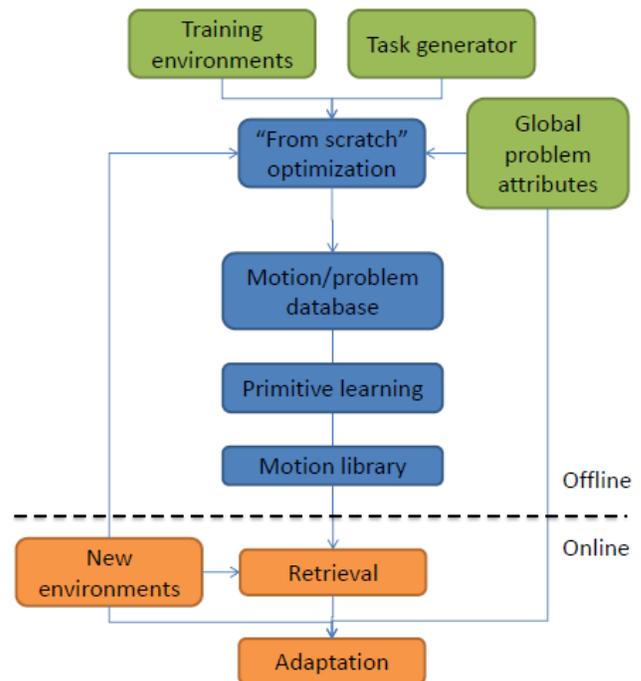

**Figure 1**. An illustration of the motion library approach.

If the motion library were sufficiently rich and retrieval were sufficiently fast, the benefits to such a scheme would be clear: *robots would respond faster*, because extensive optimization on-line would be avoided; *robots would be able to execute unintuitive behaviors at their performance limits*, because optimization is not limited to an engineer's imagination; and *robots would be more capable*, because an essentially infinite number of problem variations can be explored in simulation. Skills will no longer need to be painstakingly-crafted in the lab; an engineer will simply need to provide additional test environments and wait for a modest amount of precomputation time before a new skill emerges automatically.

This paper presents a vision for the new motion library framework and discusses promising research directions for making it feasible. It remains unresolved whether the approach is computationally feasible, whether libraries can be sufficiently rich to cover all problems of interest, and whether retrieval can be made sufficiently fast. But we observe broader computing trends that give us reason to be optimistic. First, library precomputation is trivially parallelizable, and costs are rapidly dropping as vast amounts of computing resources are becoming readily available via high-performance clusters and cloud computing. I argue that with the right computing infrastructure, it would be orders of magnitude cheaper and faster to calculate robot behaviors automatically than to employ human labor to develop them. Second, information retrieval techniques for documents, images, and 3D objects can access relevant queries from databases containing billions of entries in a fraction of a second. Extending them to handle problems and motions will require a great deal of new work, but the challenge is by no means insurmountable.

## II. BACKGROUND AND PRIOR WORK

The idea of robot learning is appealing, and has been studied in past work in many forms such as reinforcement learning [1], iterative learning control [2], and dynamic movement primitives [3]. However, knowing "when to use" a motion strategy is still a challenge, because rather than learning in the space of states, it requires learning in the space of problems, which is infinite-dimensional. Hence, learning from physical experience or manual teaching typically fails to provide sufficiently large training sets to select appropriate strategies.

Motion libraries have been studied most significantly in the virtual character animation community. Several techniques exist for generating novel motions from high-quality human motion clips, either by sequencing several motions [4,5] or adapting motions to new characters [6,7]. It is also possible to learn a probability distribution of natural-looking poses from human motion capture data, and to bias the solution of optimization problems toward those poses [8, 9]. The successes of this approach suggest that many complex, multi-step motions can be quickly composed of a relatively small number of simple, reusable subsegments (e.g., stepping motions). This paper outlines a similar approach, but one that does not presuppose the existence of human motion datasets. It also puts a higher priority on physical feasibility through the use of constrained global optimization.

In robotics, past efforts on optimization-based motion learning and adaptation include [10,11,12,13,14]. In general, this research has suggested that a small amount of online optimization to the novel problem (adaptation) makes it less important to learn optimal motions precisely (Figure 2). As a result they are able to use manageably small motion libraries (dozens or hundreds of primitives) to address novel problems. This paper considers the new research issues that will need to be tackled to scale up to massive numbers of primitives.

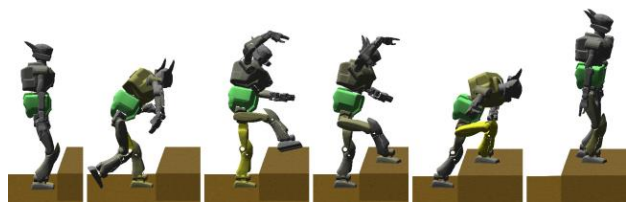

(a) Stair step planned entirely from scratch.

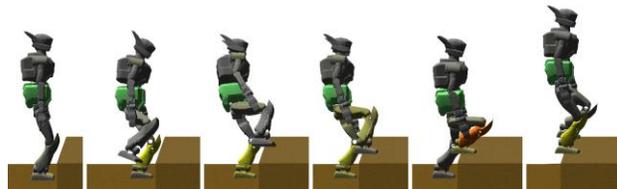

(b) Primitive adaptation leads to a more natural looking motion.

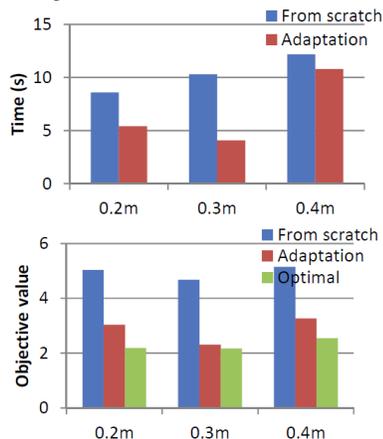

(c) Adaptation planning can produce high quality paths in less time than planning from scratch (lower objective values are better).

**Figure 2**. Adaptation planning for a humanoid robot (reprinted from [14]).

## III. MOTION LIBRARY APPROACH

This section describes a high level overview of the approach and why it is likely to be economically viable.

### A. The Motion Library Workflow

Unlike the current state of practice of dedicating tens or hundreds of thousands of man-hours toward engineering robot behaviors, the motion library approach considers the following workflow:

- A robot model, a set of representative training environments, and a task generator are sent to a computing server (e.g., the cloud).

- The server precomputes a massive motion library, including similarity metrics and data structures for optimized retrieval of appropriate motion primitives.

- Either (A) the motion library is transferred to physical robots for local retrieval, or (B) the robots query the motion library remotely from the server.

Periodically, the server may continue to expand the motion library to incrementally improve the robots' performance. Robots may provide feedback about their deployed environments, which helps the motion library adapt over time.

## B. Mathematical Formulation

The mathematical formulation of the motion library approach is highly general and straightforward [10]. If $p$ is a problem specification, $x$ is a candidate motion, and $f(x;p)$ is a quality metric (higher is better), we wish to learn an approximation of the map from problems to optimal motions:

$$g(p) \approx x^* \equiv \arg\max_x f(x;p). \tag{1}$$

A motion library is a set of $N$ problem/motion pairs $(p[1],x[1]),...,(p[N],x[N])$. Problems include environmental variables (which are typically complex, e.g., 3D maps) and task variables (e.g., start and goal configuration). To produce the large amounts of training data needed to make learning work, a relatively small number of *manually-provided training environments* is combined with a *task generator*, which samples task variables at random according to a given task distribution.

Given a problem similarity metric $d$, we can define the primitive retrieval function:

$$retrieve(p) = x[index(p)] \tag{2}$$

where $index(p) = \arg\min_k d(p[k],p)$ is the index of the nearest problem to $p$. Let us put aside for the moment the issue of defining problem similarity; this issue will be revisited in Section IV.

Now, the final adaptation layer is represented via a map $adapt(x,p)$ from a "guess" $x$ to a solution of problem $p$. In the simplest unconstrained case, $adapt(x,p)=x$, but more typically the constraints in $p$ will need to be taken into account. The final representation of $g(p)$ is therefore

$$g(p) = adapt(retrieve(p),p). \tag{3}$$

This can be considered a form of nonparametric estimator of the ideal $g$.

The designer of a motion library must carefully consider the time-quality tradeoff when designing the library size, *adapt* routine, and *retrieve* routine. With respect to solution quality, the method is expected to perform better with $N$ large and strong adaptation, i.e., is likely to map even poor guesses to high quality solutions. With respect to time, this system will perform better when $N$ is low or fast indexing structures help compute (2) quickly, and weak adaptation.

## C. The Economic Calculus

Developing planning and control strategies for robots is labor-intensive. Consider that in 2009, Willow Garage developed behaviors for the PR2 household robot to open doors and plug itself in for recharging batteries [15]. A human labor cost from ~$50,000-400,000 per robot behavior is estimated given the following assumptions:

- 1-4 engineers at 6-12 months development time per behavior

- $100,000 annual salary (reasonable for software engineers in Silicon Valley).

On the other hand, with Amazon EC2's cloud computing service, a machine can be rented on-demand for $0.06/hour [16]. So, $100,000 buys 1.6 million hours of computation, taking approximately one month of work from 4,000 machines. A motion library approach could compute 70 behaviors for the same cost as an engineer's yearly salary, and at a much faster rate, given the (very rough) assumptions:

- One robust behavior consists of 1,000 optimized motions (e.g., a motion library of size 1,000 successfully solves all problem variations expected to be solved by the behavior)

- A single machine optimizes 1 motion / day.

Granted, these estimates should be taken with a grain of salt. The conversion rate between a "robust behavior" and optimized motion is completely unclear, and the motion optimization speed depends on the complexity of the robot and environment. Furthermore, human labor will still be needed to configure the precomputations by providing test problems, simulations, and problem features, and to verify and test behaviors on the robot. Nevertheless, these calculations suggest that motion libraries have the potential to rapidly accelerate the development of robot behaviors without increasing costs.

## IV. New Enabling Technologies

A good motion library will exhibit *fast retrieval* and *wide applicability* while relying only on *weak adaptation*, because online costs are minimized while maintaining high solution quality. Tools from data mining and information retrieval may be useful to learn good libraries from a huge number of raw input motions.

## A. Clustering, Problem Features, and Indexing

First, clustering and segmentation may be used to reduce a huge motion space $N$ into a manageable number of motion primitives. Second, due to the complexity of 3D environments, problems will likely need to be indexed by a feature vector rather than a direct representation. It is unclear which of a large number of features may be most predictive of accurate retrievals, but unsupervised feature selection techniques may be useful. Third, approximate nearest neighbors or locality sensitive hashing techniques can be used to achieve sub-linear lookup times even with high $N$.

## B. Adaptation-Sensitive Problem Similarity Metrics

Let us now revisit the issue of problem similarity metrics. Suppose for sake of argument that we could test the adaptation quality for *every* primitive. Then, we would see that the optimal retrieval index $index^*(p)$ is:

$$index^*(p) = \arg\max_k f(adapt(x[k],p),p). \tag{4}$$

Hence, an *optimal* similarity metric $d$ will be one for which arg min$_k$ $d(p[k],p)$=*index*\*$(p)$ holds over all problems. Such a metric is *adaptation-sensitive* because it depends directly on the adaptation process. Of course, we cannot determine *index*\*$(p)$ without applying *adapt* to all primitives, which would largely defeat the purpose of a motion library.

However, we can learn a problem-space metric that *approximates* $d(p[k],p) \propto$ -$f(adapt(x[k],p),p)$. Such a Quality-of Adaptation (QoA) metric would result in a close approximation *index*$(p) \approx$ *index*\*$(p)$, and hence, high quality adaptations. It is also a simple matter to consider computational costs in the measure of adaptation quality. To train a QoA metric, we may select a sample of source/target (s,t) pairs from the motion library and compute training examples $d(p[s],p[t])$ = -$f(adapt(x[s],p[t]),p[t])$. Any supervised method then can be used to learn the function $d(p,p')$.

## V. CONCLUSION AND VISION FOR FUTURE WORK

This paper outlined a vision for generating and using motion libraries of unprecedented scale to solve the real-time global optimization problems that are ubiquitous in robotics. The approach is outlined in a mathematically sound framework, and is argued to be economically viable compared to human labor in generating robust robot behaviors.

Future research should address whether a library can represent repeatable motion patterns:

1. *How large must a motion library be to tackle complex problems*? Existing techniques can handle dozens of primitives, but new retrieval techniques are needed to scale to thousands or millions. If billions or trillions of motions are needed, then the approach is likely impractical.

2. *How can common motifs* (steering maneuvers, footsteps, grasps) *be clustered and segmented* to be used as primitives in the vast amounts of generated data?

And how to implement primitive retrieval:

3. *What problem features and indexing structures yield fast and effective primitive retrieval*? Image retrieval techniques scale to millions of images due to decades of research in high-quality image features (e.g., SIFT, HOG) and approximate nearest neighbors techniques, while research on robot control problem retrieval is practically nonexistent.

4. *How does the power of the adaptation routine affect library applicability and responsiveness*? Stronger adaptation lessens the need for larger libraries, at the expense of more online computation.

This paper outlined a number of promising approaches for addressing these research challenges, and sketched out the idea of learning adaptation-sensitive problem similarity metrics.

Outside of the scope of this paper, but still important for the success of a motion library method, are research directions in closing the loop with high-level planning and perception:

5. How can planners efficiently compose long-term, high-level behavior out of primitives, particularly where multiple primitives appear equally favorable to achieve a goal?

6. How should the robot incorporate sensing feedback to compensate for simulation errors?

Should this effort be successful in overcoming the intractability of global optimization, it could usher in dramatic advances in real-time control of complex tasks, such as those that are typical in household, industrial, and space robots.

## REFERENCES

[1] L.P. Kaelbling, M.L. Littman, A.W. Moore. *Reinforcement learning: A survey*. Journal of Artificial Intelligence Research, Vol 4, pages 237-285, 1996.

[2] K.L. Moore. *Iterative learning control: an expository overview*. Applied and computational control, signals, and circuits, 1999.

[3] S. Schaal, J. Peters, J. Nakanishi, and A. Ijspeert. *Learning movement primitives*. Robotics Research, Springer Tracts in Advanced Robotics, vol 15, pages 561-572, 2005.

[4] Okan Arikan and David A. Forsyth. *Interactive motion generation from examples*. ACM Transactions on Graphics (ACM SIGGRAPH 2002), 21(3):483-490, 2002.

[5] Lucas Kovar, Michael Gleicher, and Frederic Pighin. *Motion graphs*. In SIGGRAPH, pages 473-482, San Antonio, Texas, 2002.

[6] Michael Gleicher. *Retargetting motion to new characters*. In SIGGRAPH, pages 33-42, 1998.

[7] Zoran Popovic and Andrew Witkin. Physically based motion transformation. In SIGGRAPH, pages 11-20, 1999.

[8] Keith Grochow, Steven L. Martin, Aaron Hertzmann, and Zoran Popovic. *Style-based inverse kinematics*. ACM Trans. Graph., 23(3):522-531, 2004.

[9] Katsu Yamane, James J. Kuffner, and Jessica K. Hodkins. *Synthesizing animations of human manipulation tasks*. ACM Trans. Graph., 23(3):532¦539, 2004.

[10] Nikolay Jetchev and Marc Toussaint. *Trajectory prediction: learning to map situations to robot trajectories*. Proceedings of the 26th International Conference on Machine Learning (ICML-09), 2009.

[11] Nikolay Jetchev and Marc Toussaint. *Fast motion planning from experience: trajectory prediction for speeding up movement generation*. Autonomous Robots. 34(1-2): 111-127, January 2013.

[12] A. Dragan, G. Gordon, and S. Srinivasa. *Learning from Experience in Manipulation Planning: Setting the Right Goals*. Int'l Symposium on Robotics Research, 2011.

[13] Berenson, D.; Abbeel, P.; Goldberg, K. *A robot path planning framework that learns from experience*. IEEE Int'l Conference on Robotics and Automation (ICRA), May 2012.

[14] Kris Hauser, Timothy Bretl, Kensuke Harada, and Jean-Claude Latombe. *Using motion primitives in probabilistic sample-based planning for humanoid robots*. In WAFR, New York, NY, 2006.

[15] J. Markoff. *Opening Doors on the Way to a Personal Robot*. New York Times, June 8, 2009.

[16] Amazon EC2 Pricing. http://aws.amazon.com/ec2/pricing/ (retrieved on 5/1/2013).

# Predicting the Change – A Step Towards Life-Long Operation in Everyday Environments

Niko Sünderhauf, Peer Neubert, Peter Protzel

Department of Electrical Engineering and Information Technology
Chemnitz University of Technology, Germany
{niko.suenderhauf, peer.neubert, peter.protzel}@etit.tu-chemnitz.de

*Abstract*—Changing environments pose a serious problem to current robotic systems aiming at long term operation. While place recognition systems perform reasonably well in static or low-dynamic environments, severe appearance changes that occur between day and night, between different seasons or different local weather conditions remain a challenge. In this paper we propose to learn to *predict* the changes in an environment. Our key insight is that the occurring scene changes are in part systematic, repeatable and therefore predictable. The goal of our work is to support existing approaches to place recognition by learning how the visual appearance of an environment changes over time and by using this learned knowledge to predict its appearance under different environmental conditions. We describe the general novel idea of scene change prediction and a proof of concept implementation based on vocabularies of superpixels. We can show that the proposed approach improves the performance of SeqSLAM and BRIEF-Gist for place recognition on a large-scale dataset that traverses an environment under extremely different conditions in winter and summer.

## I. Introduction

Long term operation in changing environments is one of the major challenges in robotics today. Robots operating autonomously over the course of days, weeks, and months are faced with significant changes in the appearance of an environment: A single place can look extremely different depending on the current season, weather conditions or the time of day. Since state of the art algorithms for autonomous navigation are often based on vision and rely on the system's capability to recognize known places, such changes in the appearance pose a severe challenge for any robotic system aiming at autonomous long term operation.

The problem has recently been addressed by few authors, but so far no congruent solution has been proposed. Milford and Wyeth [3] proposed to increase the place recognition robustness by matching *sequences* of images instead of single images and achieved impressive results on two across-seasons datasets. Exploring into a different direction, Churchill and Newman [2] proposed to accept that a single place can have a variety of appearances. Their conclusion was that instead of attempting to match different appearances across seasons or severe weather changes, different *experiences* should be remembered for each place, where each experience covers exactly one appearance. Both suggested approaches can be understood as the extreme ends of a spectrum of approaches that spans between interpreting changes as individual experiences of a single place on one hand and increasing the robustness of

**State of the Art Approach**

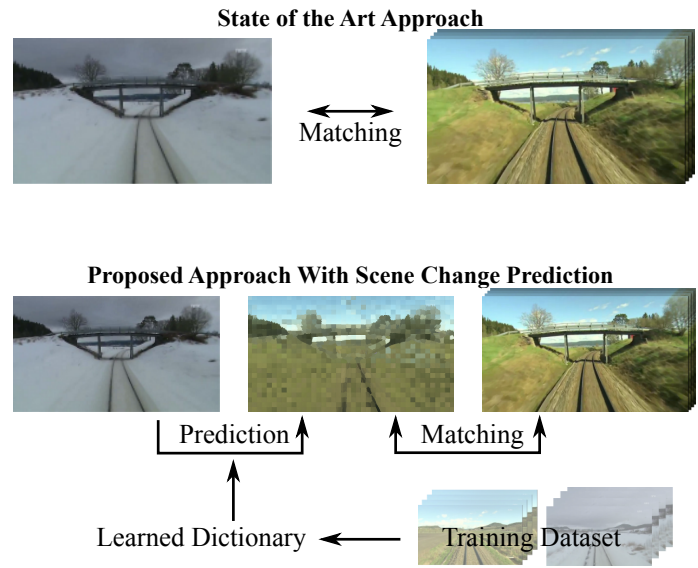

**Proposed Approach With Scene Change Prediction**

Fig. 1. State of the art approaches to place recognition attempt to directly match two scenes, even if they have been observed under extremely different environmental conditions. This is prone to error and leads to bad recognition results. Instead, we propose to *predict* how the query scene (the winter image) would appear under the same environmental conditions as the database images (summer). This prediction process uses a dictionary that exploits the systematic nature of the seasonal changes and is learned from training data.

the matching against appearance changes on the other hand. Our work presented in the following is orthogonal to this spectrum.

What current approaches to place recognition (and environmental perception in general) lack, is the ability to *reason* about the occurring changes in the environment. Most approaches try to merely *cope* with them by developing change-invariant descriptors or matching methods. Potentially more promising is to develop a system that can *learn* to *predict* certain systematic changes (e.g. day-night cycles, weather and seasonal effects, re-occurring patterns in environments where robots interact with humans) and to infer further information from these changes. Doing so without being forced to explicitly know about the *semantics* of objects in the environment is in the focus of our research and the topic of this paper.

Fig. 1 illustrates the core idea of the paper and how it compares to the current state of the art place recognition algorithms. Suppose a robot re-visits a place under extremely different environmental conditions. For example, an environment was first experienced in summer and is later re-visited in winter time. Most certainly, the visual appearance has undergone extreme changes. Despite that, state of the art approaches would attempt to match the currently seen winter image against the stored summer images.

Instead, we propose to *predict* or *hallucinate* how the current scene would appear under the same environmental conditions as the stored past representations, before attempting to match against the database. That is, when we attempt to match against a database of summer images but are in winter time now, we predict how the currently observed winter scene would appear in summer time or vice versa.

The result of this prediction process depends on the actual place recognition algorithm that is applied. When using approaches like SeqSLAM [3] or BRIEF-Gist [4], the result would be a synthesized *image* as illustrated in Fig. 1. This image preserves the structure of the original scene but is close in visual appearance to the corresponding original summer scene. When using place recognition based on a bag of words approach (e.g. FAB-MAP), the result of the prediction process would be a translated bag of words.

In any case, the proposed prediction can be understood as *translating* the image from a winter vocabulary into a summer vocabulary or from winter language into summer language. As is the case with translations of speech or written text, some details will be lost in the process, but the overall *idea*, i.e. the gist of the scene will be preserved. Sticking to the analogy, the error rate of a translator will drop with experience. The same can be expected of our proposed system: It is dependent on training data, and the more and the better training data it gets, the better can it learn to predict how a scene changes over time or even across seasons.

To the best of our knowledge, the idea of predicting extreme scene changes across seasons to aid place recognition is novel and has not been proposed before.

## II. LEARNING TO PREDICT SCENE CHANGES ACROSS SEASONS

How can the severe changes in appearance a landscape undergoes between winter and summer be learned and predicted? The underlying idea of our approach is that the appearance change of the whole image is the result of the appearance change of its parts. If we had an idea of the behavior of each part, we could predict the whole image. However, instead of trying to recover semantic information about the image parts and model their behavior explicitly, we make the assumption that similarly *appearing* parts change their appearance in a similar way. While this is for sure not always true, it seems to hold for many practical situations. This idea can be extended to groups of parts, incorporating their mutual relationships.

To predict how the appearance of a scene changes between different conditions (e.g. summer and winter), we propose to first conduct a learning phase on training data. This data comprises scenes observed under both summer and winter conditions. In the subsequent prediction phase, the change in appearance of a new scene can be predicted using the results of the training phase. In the following, we explain our current proof of concept implementation of the proposed scene change prediction approach.

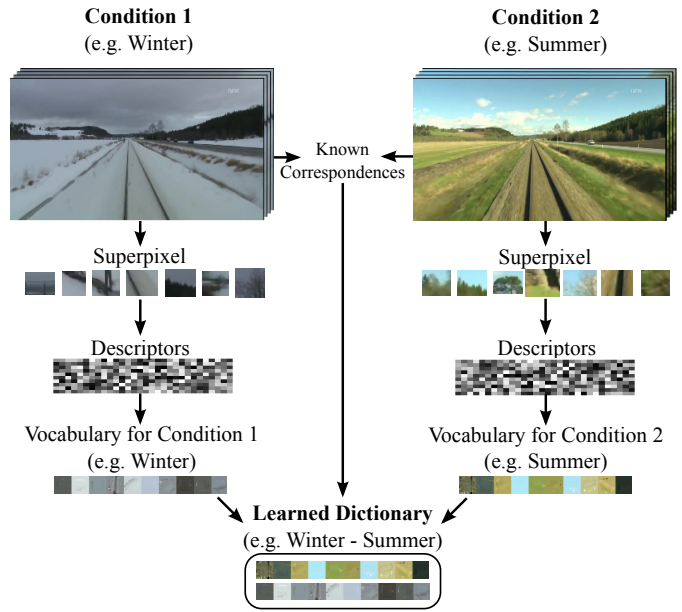

Fig. 2. Learning a dictionary between images under different environmental conditions (e.g. winter and summer). The images are first segmented into superpixels and a descriptor is calculated for each superpixel. These descriptors are then clustered to obtain a vocabulary of visual words for each condition. In a final step, the dictionary that translates between both vocabularies is learned. This can be done due to the known pixel-accurate correspondences between the input images.

### A. Learning Vocabularies and a Dictionary

During the training phase we have to learn a vocabulary for each viewing condition and a dictionary to translate between them. In a scenario with two viewing conditions (e.g. summer and winter), the input to the training are images of the same scenes under both viewing conditions and known associations between pixels corresponding to the same world point. Obviously the best case would be perfectly aligned pairs of images, e.g. captured by stationary webcams. Which approach to visual vocabulary learning is the most promising for the proposed scene change prediction has to be evaluated in future work.

Fig. 2 illustrates the training phase. In our current proof of concept implementation, each image is segmented into SLIC superpixels [1]. For each superpixel a descriptor that contains a color histogram in LAB color space (each channel with 10 bins), an U-SURF descriptor (128 byte) to capture texture information and the $y$-coordinate to encode spatial information is computed. The set of descriptors for each viewing condition is clustered to a vocabulary using hierarchical k-means. Each cluster center becomes a word in this visual vocabulary. The descriptors and the average appearance of each word (the word patch) are stored for later synthesizing of new images. For our experiments, we learned 10.000 words for each vocabulary.

Given the learned vocabularies, we can proceed to learn

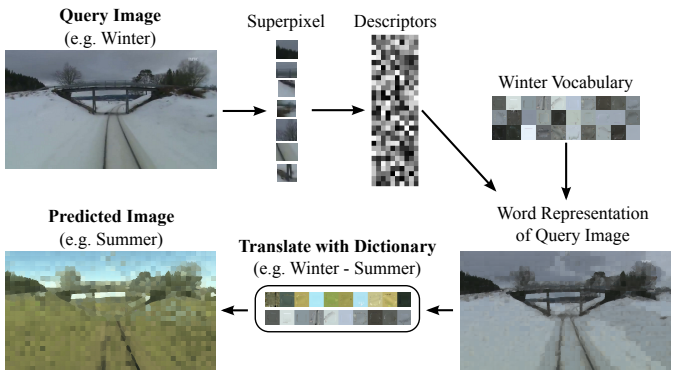

**Query Image**
(e.g. Winter)

Superpixel

Descriptors

Winter Vocabulary

Word Representation
of Query Image

**Predicted Image**
(e.g. Summer)

**Translate with Dictionary**
(e.g. Winter - Summer)

Fig. 3. Predicting the appearance of a query image under different environmental conditions: How would the current winter scene appear in summer? The query image is first segmented into superpixels and a descriptor is calculated for each of these segments. With this descriptor each superpixel can be classified as one of the visual words from the vocabulary. This word image representation can then be translated into the vocabulary of the target scene (e.g. summer) through the dictionary learned during the training phase. The result of the process is a synthesized image that predicts the appearance of the winter query image in summer time.

a dictionary that can translate between visual words from two environmental "languages" or conditions as illustrated in the lower part of Fig. 2. This dictionary captures the transitions of the visual words when the environmental conditions change. The dictionary can either capture the complete discrete probability distribution of these transitions or only store the transition that occurs most often.

### B. Predicting Image Appearances Across Seasons

Fig. 3 illustrates how we can use the learned vocabularies and the dictionary to predict the appearance of a query image across different environmental conditions.

The query image is segmented into superpixels and a descriptor for each superpixel is computed. Using this descriptor, a word from the vocabulary corresponding to the current environmental conditions (e.g. winter) is assigned to each superpixel. The learned dictionary between the query conditions and the target conditions (e.g. winter-summer) is used to translate these words into words of the target vocabulary.

If the vocabularies also contain *word patches*, i.e. an expected appearance of each word, we can synthesize the predicted image based on the word associations from the dictionary and the spatial support given by the superpixel segmentation.

## III. EXPERIMENTS AND RESULTS

After the previous section explained how scene change prediction across seasons can be performed, we are going to describe the conducted experiments and their results.

### A. The Nordland Dataset

To test our proposed approach of scene change prediction, we required a dataset where a camera traverses the same places under very different environmental conditions but under a similar viewing perspective: The TV documentary "Norlandsbanen – Minutt for Minutt" by the Norwegian Broadcasting

Corporation NRK provides video footage of a 728 km long train ride that has been filmed from the perspective of the train driver four times in spring, summer, fall, and winter. The full-HD recordings have been time-synchronized such that an arbitrary frame from one video corresponds to the same frame of any of the other three videos etc. Therefore, frame-accurate ground truth information, e.g. corresponding scenes, are available. Furthermore, since the cameras were mounted exactly in the same spot in the driver's cabin, the four videos are almost perfectly aligned and thus allow easy learning of visual word transitions between the four seasons. The videos are available online at http://nrkbeta.no/2013/01/15/nordlandsbanen-minute-by-minute-season-by-season/ under a Creative Commons licence (CC-BY).

For our experiments described in the following we extracted 30 minutes from the spring and the winter videos, starting approximately at 2 hours into the drive. From the four available videos, the spring video best resembled typical summer weather conditions. To form the training dataset, we extracted approximately 900 frames from the first 8 minutes of this 30 minutes subset. This training dataset was used to learn the visual vocabulary for summer and winter and the dictionary to translate between both seasons. The remaining 22 minutes of the video subset served as the test dataset to evaluate the performance of the proposed scene change prediction.

### B. Extending and Improving BRIEF-Gist and SeqSLAM

SeqSLAM [3] and BRIEF-Gist [4] are two established approaches to appearance-based place recognition. BRIEF-Gist is a holistic descriptor that encodes the visual appearance of a whole image in a short bit string. It supports place recognition by applying the Hamming distance between two descriptors in order to find the single global best matching query image. In contrast, SeqSLAM performs place recognition by matching whole *sequences* of images and has been shown to perform well despite severe appearance changes [3, 5]. We use OpenSeqSLAM [5] to perform the experiments.

Combining both approaches with our scene change prediction is particularly easy, since the change prediction algorithm can be executed as a preprocessing step before SeqSLAM or BRIEF-Gist start with their own processing. Since we attempted to match summer against winter images, we predicted the visual appearance of each summer scene in winter and fed the predicted winter images together with the original real winter images into BRIEF-Gist and SeqSLAM.

Fig. 4 compares precision-recall curves achieved by both algorithms with and without our proposed scene change prediction. The apparent result is that both BRIEF-Gist and SeqSLAM can immediately benefit from the change prediction. For SeqSLAM we plot the results for several values of the `ds` parameter that controls the minimal required length of the matched image sequences in seconds. We can see that SeqSLAM's performance increases with larger `ds`, as expected.

We can conclude that although SeqSLAM alone reaches good matching results, it can be significantly improved by first predicting the appearance of the query scene under the viewing conditions of the stored database scenes. Also BRIEF-Gist can benefit from the proposed appearance change prediction, although its performance is in general much worse than SeqSLAM's.

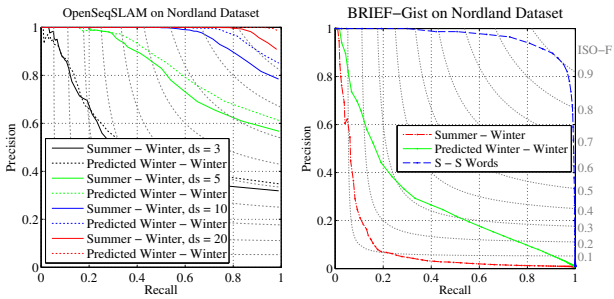

Fig. 4. Precision recall plots obtained by place recognition across seasons with SeqSLAM [3] (left) and BRIEF-Gist [4] (right). The plots compare the performance of the stand-alone algorithms with the boosted performance when the appearance of the winter images is predicted before place recognition is attempted. It is apparent that our proposed approach can significantly improve the performance of both algorithms. For comparison, the blue curve in the right plot shows the performance of BRIEF-Gist when matching summer images directly with summer word images, i.e. performing place recognition under constant environmental conditions.

For comparison we also evaluated the performance of FAB-MAP (using openFAB-MAP) on the dataset. As expected, directly matching winter against summer images was not successful. The best measured recall was 0.025 at 0.08 precision, presumably because FAB-MAP fails to detect common features in the images from both seasons.

## IV. Discussion and Conclusions

Our paper described the novel concept of learning to predict systematic changes in the appearance of environments. We explained our implementation based on superpixel vocabularies and demonstrated how two approaches to place recognition, BRIEF-Gist and SeqSLAM, can benefit from the scene change prediction step.

We can synthesize an actual image during this prediction. This simplifies the qualitative evaluation by visually comparing the predicted with the real images and further allows to use existing place recognition algorithms for quantitative evaluation. However, the proposed idea of scene change prediction can in general be performed on different levels of abstraction: It could also be applied *directly* on holistic descriptors like BRIEF-Gist, on visual words like the ones used by FAB-MAP or on the downsampled and patch-normalized thumbnail images used by SeqSLAM. Furthermore, the learned dictionary can be as simple as a one-to-one association or capture a full distribution over possible translations for a specific word. In future work this distribution could also be conditioned on the state of neighboring segments, and other local and global image features and thereby incorporate mutual influences and semantic knowledge. This could be interpreted as introducing a *grammar* in addition to the vocabularies and dictionaries. How such extended statistics can be learned from training data efficiently is an interesting direction for future work.

If the dictionary does not exploit such higher level knowledge (as in the superpixel implementation introduced here) the quality of the prediction is limited. In particular, when solely relying on local appearance of image segments for prediction, the choice of the training data is crucial. It is especially important that the training set is from the same domain as the desired application, since image modalities that were not well-covered by the training data can not be correctly modelled and predicted. Exploring the requirements for the training dataset and how the learned vocabularies and dictionary can best generalize between different environments will be part of our future research.

In its current form, our algorithm requires perfectly aligned images in the training phase. This requirement is hard to fulfill and limits the available training datasets. We will explore ways to ease this requirement in future work, e.g. by anchoring the training images on stable features. Another key limitation of the system in its current form is that it requires different vocabularies for *discrete* sets of environmental conditions. While it is of course possible to create and manage a larger number of such vocabularies and the respective mutual dictionaries, a unified approach that learns and maintains a single vocabulary that captures all conditions would be more desirable. As already mentioned, the Nordland dataset provides somewhat optimal conditions (apart from the season-induced appearance changes) for place recognitions, since the camera observes the scene from almost exactly the same viewpoint in all four seasons and the variability of the scenes in terms of semantic categories is rather low. These conditions would usually not be met and we therefore prepare to evaluate the proposed approach in a more general setting using data from vehicles in urban environments and training data that has been collected from stationary webcams over the course of several months.

## References

[1] Radhakrishna Achanta, Appu Shaji, Kevin Smith, Aurelien Lucchi, Pascal Fua, and Sabine Susstrunk. SLIC Superpixels Compared to State-of-the-Art Superpixel Methods. volume 34, 2012.

[2] Winston Churchill and Paul M. Newman. Practice makes perfect? Managing and leveraging visual experiences for lifelong navigation. In *Proc. of Intl. Conf. on Robotics and Automation (ICRA)*, 2012.

[3] Michael Milford and Gordon F. Wyeth. SeqSLAM: Visual route-based navigation for sunny summer days and stormy winter nights. In *Proc. of Intl. Conf. on Robotics and Automation (ICRA)*, 2012.

[4] Niko Sünderhauf and Peter Protzel. BRIEF-Gist – Closing the Loop by Simple Means. In *Proc. of IEEE Intl. Conf. on Intelligent Robots and Systems (IROS)*, 2011.

[5] Niko Sünderhauf, Peer Neubert, and Peter Protzel. Are We There Yet? Challenging SeqSLAM on a 3000 km Journey Across All Four Seasons. In *Proc. of Workshop on Long-Term Autonomy, IEEE International Conference on Robotics and Automation (ICRA)*, 2013.

# The Social Co-Robotics Problem Space:
# Six Key Challenges

Laurel D. Riek

Department of Computer Science and Engineering
University of Notre Dame
Email: lriek@nd.edu

*Abstract*—In order to realize the long-term vision of intelligent co-robots capable of competent proxemic interaction with humans, it is important that our research community fully define the problem space and articulate its inherent challenges. In particular, we must recognize that many problems in the space may not be computable or definable, and must determine ways to address this challenge moving forward. This paper broadly sketches six key challenges in the social co-robotics problem space and suggests several paths toward solving them, such as Wizard-of-Oz constraints and problem satisfaction.

## I. Introduction

Three recent U.S. Government reports and funding initiatives in robotics - the CCC Robotics Roadmap [26], the National Intelligence Council Global Trends 2030 [5], and the National Robotics Initiative [17] - all strongly emphasize the theme that in order to realize the vision of intelligent, capable co-robots, robots must be able to operate intelligently in close proximity to (and with) humans.

The co-robotics problem domain includes both proximate and remote interaction [12], and covers a wide range of human environments. In this paper, we focus specifically on problems relating to co-robots in *human social environments* (HSE). These are any environments in which a robot operates proximately with humans. We define these robots as *social co-robots*, in that they operate in an HSE, are physically embodied, and have at least some degree of autonomy.

It is worth noting that social co-robots are not necessarily *sociable* [2] - they do not necessarily need to interact with us interpersonally. For example, a service robot that empties the dishwasher may not be sociable, but because it operates in an HSE with the aforementioned characteristics it is a social co-robot.

Fig. 1 depicts the broad application space for co-robots in human environments, and emphasizes a few application areas where social co-robots in HSEs are warranted. Exemplar areas include personal assistive robots (physical and cognitive), educational robots, robots for leisure activities, service robots, and robots in clerical domains. These application areas are not meant to be mutually exclusive, but the majority of problems in, for example, biomolecular or field robotics, are not usually social in nature.

This paper will define six challenges unique to the social co-robotics problem domain (Section II), then suggest some possible avenues to explore for addressing them (Section III).

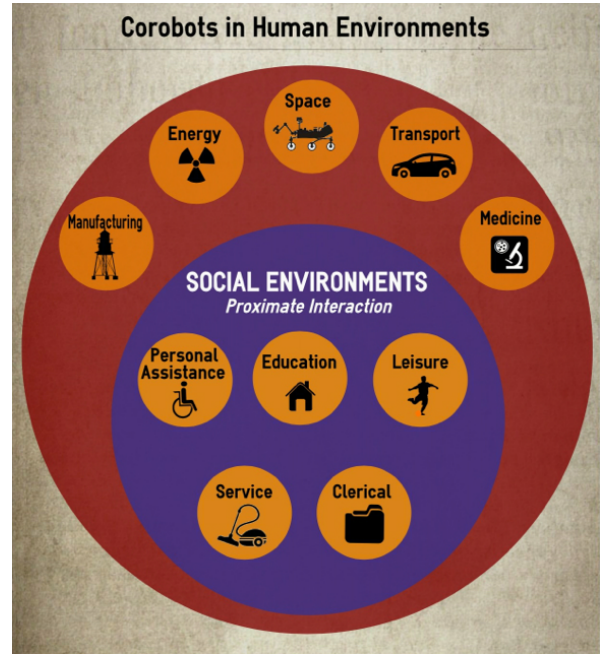

Fig. 1. The co-robotics problem domain can be divided into two domains: human social environments, where proximate interaction is required, and others, where remote interaction is more common. *None of these areas are intended to be mutually exclusive.*

## II. Six Key Challenges in Social Co-Robotics

Fig. 2 depicts several of the unique set of challenges that social co-robotics faces. While many of these problem domains are tied to the three traditional robotics problems (perception, cognition, and action), they are much more complex in scope. These problems may not be computable, or even definable. Further, several of these problems resolve into fundamental AI-complete problems, such as natural language understanding [14], making them intractable.

It is critical as a community we articulate the inherent hardness of these problems, and recognize there is no silver bullet for solving them [24].

### A. Problem 1: Dynamic Spaces

One thing unique to social co-robots is that they must operate in HSEs with humans present. Humans by their very existence create unforeseen challenges to robotics that are still relatively new to the field. A workshop parallel to this one at RSS is called "Robots in Clutter", and its CFP describes many of these challenges: vision under cases of clutter and occlusion, dynamic navigation, action planning on the fly, blind manipulation, and "sporadic user involvement" [35].

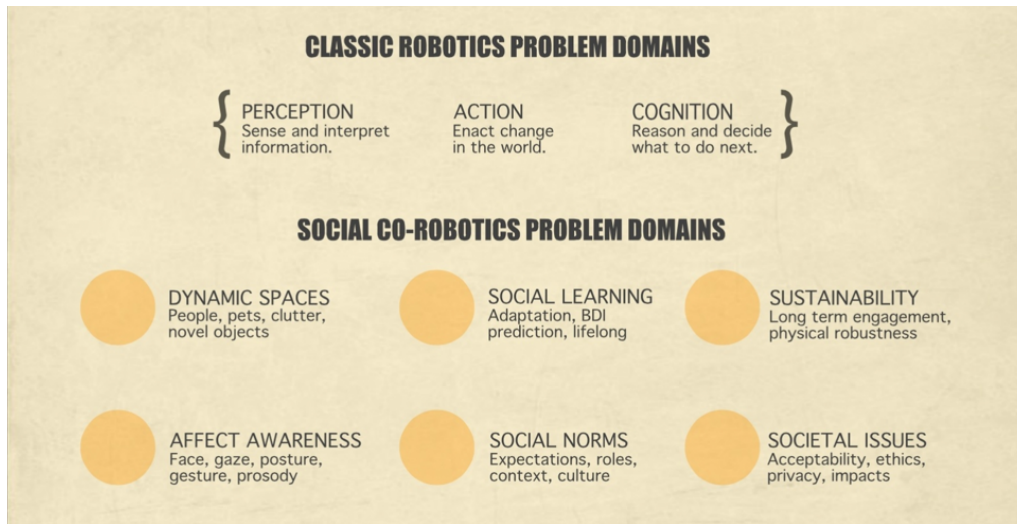

Fig. 2. Social co-robotics has a unique set of challenges, many of which are entirely unlike traditional robotics problems in terms of their level of complexity as well as the challenge in adequately defining them. Six problem domains are listed above.

HSEs are often highly fluid and variable in nature, which makes it difficult to anticipate or plan for these environmental alterations as a robot. Traditional control paradigms, even those that are well suited to and robust within dynamic environments (e.g., [3, 11]) do not scale well to these complexities.

Part of the problem is that the way in which a robot contends with a dynamic HSE is closely tied to its task and embodiment, thus making it difficult to generalize the problem. As robotics researchers, we can clearly envision a general solution to the 2005 DARPA Grand Challenge that is vehicle-independent. We cannot, however, envision a general solution to a personal assistance robot that helps a person with severe physical disabilities compete their daily tasks[1]. Solving such a problem is multifaceted, contains potentially an infinite number of dynamically changing subtasks, and any solution attempt is intricately tied to the robot's embodiment and capabilities.

Thus, this problem area may not even be definable for robotics, let alone computable. However, perhaps one way to address it is through the use of social learning, described in the next section.

### B. Problem 2: Social Learning

One of the most critical things a social co-robot needs to be able to do is learn and adapt to not only its environment, but the co-humans within it. However, because HSEs are highly fluid (as are people), it is important that social co-robots are capable of lifelong learning [31]. This learning may occur under direct human tutelage or independently.

In 2010, Carlson et al. [6] introduced NELL, the Never Ending Language Learner. This project is teaching a machine to "read the web", by extracting facts while crawling textual webpages 24 hours a day, 7 days a week. Another project, RobotsFor.Me [32], may also be suitable to help enable lifelong learning by letting people remotely log into a PR2 24/7 and teach it in an embodied, situated way.

However, even with these projects, learning about and adapting to humans, and anticipating and modeling their BDIs (beliefs, desires and intentions) is quite challenging, and suffers from the same problem as the *Dynamic Spaces* problem. Models for one paradigm with one robot with one set of capabilities (e.g., a PR2 in an office environment, with end-effectors and a Kinect) do not necessarily extend to others. BDI modeling may be computable within a virtual agent space, but may reach a point of being non-computable (or non-definable) when embodied on a co-robotic system.

### C. Problem 3: Sustainability

Another problem in social co-robotics concerns sustainability, or long-term interaction [8]. What happens when the novelty of a robot wears off? How does a robot adapt to changes in the preferences of co-humans sharing the HSE?

Many of these problems inherently require a robot to exhibit some degree of creativity, be attuned to the moods of its proximate humans, and keep an inordinate sense of time and history. However, creativity is likely NP-Complete [29], mood awareness is at best AI-Complete (see Section II-D), and the requisite granularity of storage is not easily definable - the representation strategy alone is likely AI-Complete [33].

### D. Problem 4: Affect and Social Signal Awareness

One of the more commonly explored problems in social robotics concerns the recognition and synthesis of affect [21] and social signals [34]. While the precise definition of the terms "affect" and "social signal" are frequently debated in the affective science and social computing communities, most agree that these terms denote the visual and aural channels of human communication. Thus, a socially aware machine infers meaning from a human's face, gaze, posture, gestures, proxemics, and prosody, and can generate these signals itself.

---

[1]To appreciate the deep complexity of this problem, readers are encouraged to watch this time lapse video of a person with Muscular Dystrophy performing his morning routine: http://youtu.be/aSDxVG0fVg4 [10]

There is certainly a clear need for robots operating in HSEs to be able to recognize and synthesize affect [26]. However, Picard, recognized as the founder of the field of affective computing, calls facial recognition alone one of the hardest, most complex problems in computer science [22]. Indeed, if one considers all of visual communication to be a component of natural language (as most do [30]), the problem of affect recognition is at best an AI-Complete problem: it inherently requires natural language understanding [14].

The problem of affect synthesis is unfortunately no better off from a complexity standpoint. Even in an asymmetrical dialogue, visual communication is intimately tied to dialogue content [1], situational context, *a priori* knowledge, expected norms, and, of course, *Social Learning*. Thus, this problem, too becomes at best an AI-complete problem, but at worst is not even definable let alone computable.

### E. Problem 5: Social Norms

Social norms are "a standard, customary, or ideal form of behavior to which individuals in a social group try to conform." [4]. In the social co-robotics problem domain, this encompasses several things. Social norms place constraints on a robot's actions, in that they must conform to people's expectations and the situation they are in [15]. This is not to say co-humans will not forgive robot mistakes, our own research suggests people may be willing to overlook a robot's social missteps [25]. Nonetheless, from a technology acceptance perspective, there is motivation to program robots to be aware of social norms.

The *Social Norm* awareness problem is perhaps a superset of the *Affect and Social Signal Awareness* problem, because it requires additional knowledge to contextualize and classify observed human behavior. A person screaming alone in one's house is quite different than screaming while attending a sporting event. This problem space, too, is infinite, and thus not easily definable or computable.

### F. Problem 6: Societal Issues

As social co-robots share HSEs with people, they inherently raise a plethora of societal issues, including privacy, security, acceptability, and so on. While these issues are not necessarily unique to social co-robotics, some of them may raise an alarm of intrusiveness other co-robotics domains do not need to face. To again draw on the example of biomolecular robotics; while a person may express concerns over the use of micro-scale drug delivery robots on an abstract level, they are unlikely to experience the same vitriolic response as they might to an embodied robot with agency in their home.

In his recent book *Robot Futures* [18], Nourbakhsh describes some of these vitriolic responses to robots, even in places as innocuous as a science museum. This use, misuse, and abuse of automated agents is not new, it dates back at least 60 years to Asimov's writings [16]. It has recently garnered attention in the HCI community by Parasuraman and Riley [20], who stress the importance of closely examining these attitudes and incorporating them into the design process.

This problem places a burden on social roboticists, because in addition to contending with the plethora of computational challenges that face our field, we further need to be concerned about public opinion of our robots if we ever want them to be purchased and used. Thus, iterative design and technology acceptance is critical, even at early stages of research.

## III. Paths Forward

In social co-robotics, we tend to address these monumental challenges in three ways: we ignore them, we "wizard away" the problem by having a human compute the solution, or we severely constrain the problem space.

As a community, ignoring these problems does not help us advance our discipline. Instead, we suggest judicious use of human computation (wizards) and developing new techniques for problem satisficing as paths forward. We discuss these in further detail below.

### A. Judicious Use of Wizard-of-Oz

Presently, many researchers in social co-robotics use Wizard of Oz (WoZ) control extensively, to solve most of the aforementioned problems, such as natural language processing, social understanding, dynamic space operation, etc. [23]. The original idea behind the WoZ paradigm was to be part of an iterative design process, a small aid in development as other components came to fruition [13]. The paradigm was not intended to be an end of and in itself; however, it has lately been re-tasked to enable robotics researchers to "project into the future" [8], enabling experiments which would be impossible with present technology.

While at first is a compelling idea, at closer inspection the majority of problems a wizard is simulating in these robotic systems actually fall within the aforementioned six major problem domains of social co-robots. In other words, *wizards are simulating AI-Complete, non-definable, and non-computable problems.*

In no way do we argue for the total abdication of the WoZ paradigm from social co-robotics research; instead, we suggest roboticists be more careful in how they employ it. For a roboticist intentionally designing a semi-autonomous robot that will have human help when making decisions, it may make sense to employ WoZ as a kind of real-time human computation [27]. Other areas in the field of Artificial Intelligence facing NP-hard problems have embraced this paradigm, such as image labeling and machine translation - why not social co-robots?

But for researchers using WoZ purely as a method for testing complex psychological hypotheses involving co-robot acceptance, we urge judicious use of its employment. We may never have robots capable of some of these tasks, so it may be scientifically disingenuous and unrealistic to run experiments assuming their existence.

## B. Problem Satisficing: "Good Enough" Social Co-robots

It may be the case that we cannot solve these monumental challenges facing our field; perhaps these problems will remain non-definable and non-computable forever. Nonetheless, it may be possible that we can build "good enough" social co-robots, where we solve problems well enough to enable adequate operation in HSEs.

Davis [9] (invoking Simon [28]) writes "nature is a satisficer, not an optimizer". Organisms solve problems in an acceptable or satisfactory (though not necessarily optimal) way. Why not robots? Certainly people have considered satisficing controllers for robotics problems in the past (c.f. [19], [7]), thus, we may be able to imagine ways to satisfice within the social co-robotics problem space without over-reduction.

It is not yet clear what the dimensions of such a satisfaction might look like for social co-robots in HSEs; however, this seems to be a wide open area of research. What is the minimum functionality a social co-robot needs to complete its tasks in HSEs? What level of failure are co-humans willing to tolerate and excuse? There are many interesting questions in this domain, and we look forward to exploring them.

### Acknowledgments

This material is based upon work supported by the National Science Foundation under Grant No. IIS-1253935.

### References

[1] J.B. Bavelas, L. Coates, and T Johnson. Listeners as co-narrators. *J Pers Soc Psychol*, 2000.

[2] C. Breazeal. *Designing Sociable Robots*. MIT Pr., 2004.

[3] R A Brooks. *Intelligence without representation*. MIT Press, 1991.

[4] M. Burke and H. P. Young. Norms, customs and conventions. In *Handbook of social economics*. 2010.

[5] M. Burrows. Global Trends 2030: Alt. Worlds. 2012.

[6] A Carlson, J Betteridge, B Kisiel, B Settles, E Hruschka, and T Mitchell. Toward an architecture for never-ending language learning. *AAAI*, 2010.

[7] J A Conlin. Getting around: making fast and frugal navigation decisions. *Prog Brain Res.*, 174, 2009.

[8] K. Dautenhahn. Methodology and themes of human-robot interaction: A growing research field. *International Journal of Advanced Robotic Systems*, 4(1), 2007.

[9] R. Davis. What Are Intelligence? And Why? *American Association for Artificial Intelligence*, 1998.

[10] ExtraAmpersand. My Disability: a real look at my life with Becker's MD. URL http://youtu.be/aSDxVG0fVg4.

[11] D. Fox, W. Burgard, and S. Thrun. The dynamic window approach to collision avoidance. *IEEE Robotics and Automation*, 4(1):23–33, 1997.

[12] M. A Goodrich and A. C Schultz. Human-robot interaction: a survey. *Foundations and Trends in Human-Computer Interaction*, 1(3):203–275, 2007.

[13] T. D Kelley and L. N Long. Deep Blue Cannot Play Checkers: The Need for Generalized Intelligence for Mobile Robots. *Journal of Robotics*, (21), 2010.

[14] L. Lee. I'm sorry Dave, I'm afraid I can't do that": Linguistics, Statistics, and Natural Language Processing. *Computer Science: Reflections on the Field*, 2004.

[15] M. Lohse. *Investigating the influence of situations and expectations on user behavior: empirical analyses in human-robot interaction*. PhD thesis, Bielefeld University, 2010.

[16] L. McCauley. Countering the Frankenstein Complex. *American Association for Artificial Intelligence*, 2007.

[17] National Science Foundation. National Robotics Initiative. URL http://nsf.gov/nri.

[18] I. Nourbakhsh. *Robot Futures*. MIT Pr., 2013.

[19] T. J. Palmer and M. A. Goodrich. Satisficing anytime action search for behavior-based voting. 1, 2002.

[20] R. Parasuraman and V Riley. Humans and automation: Use, misuse, disuse, abuse. *Human Factors*, 1997.

[21] R. W. Picard. *Affective computing*. MIT Press, 1997.

[22] R W Picard. Emotion Technology: From Autism to Customer Experience and Decision Making. *Microsoft Research Cambridge*, 2009.

[23] L. D. Riek. Wizard of Oz Studies in HRI: A Systematic Review and New Reporting Guidelines. *Journal of Human Robot Interaction*, 1(1):119–136, 2012.

[24] L. D. Riek and P. Robinson. Challenges and Opportunities in Building Socially Intelligent Machines. *IEEE Signal Processing*, 28(3):146–149, 2011.

[25] L. D. Riek, T. Rabinowitch, P. Bremner, A.G. Pipe, M. Fraser, and P. Robinson. Cooperative gestures: Effective signaling for humanoid robots. *5th ACM/IEEE Int'l Conference on Human-Robot Interaction (HRI)*, 2010.

[26] Robotics-VO. A Roadmap for U.S. Robotics: From Internet to Robotics. Technical report, 2013.

[27] D Shahaf and E Amir. Towards a Theory of AI Completeness. *8th International Symposium on Logical Formalizations of Commonsense Reasoning*, 2007.

[28] H. A. Simon. Rational choice and the structure of the environment. *Psychological Review*, 63(2), 1956.

[29] William Squires. Creative computers: Premises and promises. *Art Education*, 36(3):21–23, 1983.

[30] J Streeck. Gesture as communication I: Its coordination with gaze and speech. *Communication Monographs*, 60 (4), 1993.

[31] S. Thrun and T. Mitchell. Lifelong robot learning. *Robotics and Autonomous Systems*, 15, 1995.

[32] R Toris and S Chernova. RobotsFor.Me and Robots For You. *IUI Interactive Machine Learning Workshop*, 2013.

[33] M C Torrance and L A Stein. Communicating with martians (and robots). Technical report, 1997.

[34] A. Vinciarelli, M Pantic, and H. Bourlard. Social signal processing: Survey of an emerging domain. *Image and Vision Computing*, 27(12):1743–1759, 2009.

[35] M Zillich, M. Bennewitz, M. Fox, J. Piater, and D. Pangercic. 2nd Workshop on Robots in Clutter: Preparing robots for the real world, June 2013. URL http://workshops.acin.tuwien.ac.at/clutter2013/.

# Wearable vision systems for personal guidance and enhanced assistance

G. López-Nicolás, A. Aladrén, and J. J. Guerrero

Instituto de Investigación en Ingeniería de Aragón - Universidad de Zaragoza, Spain

*Abstract*—The challenge addressed in this paper involves research of computer vision and robotic techniques to be part of a personal assistance system based on visual information. Main features envisioned include not only navigation assistance in known or unknown environments, but also enhanced capabilities in safety and human perception augmentation. Current systems in the context of assistive devices do not provide enough level of performance, and their capabilities are still quite limited. These are normally based on conventional sensors and, although perception can be greatly improved by sensor fusion, long term research should also consider unconventional vision devices to expand the system's potential. We refer as unconventional vision systems to those consisting in more than a classical or standard camera. So, the visual assistant will be wearable including conventional and unconventional camera systems. We aim for a human-centered system, which complements rather than replaces human abilities, allowing enhanced interaction and integration of the user in the environment. We present several complementary approaches based on different unconventional camera systems. Possible users of the investigated technologies will range from visually impaired people to users with normal visual capabilities performing specific tasks, passing also into humanoid robots.

## I. Introduction

The ability of navigating effectively in the environment is natural for people, but not easy to achieve under certain circumstances, such as the case of visually impaired people or unknown and intricate environments. In a similar way, the recognition of places, objects or signs is another fundamental ability in our daily life. Humans solve these problems mainly with vision and memory combined with our learning ability. According to the World Health Organization (WHO), in 2020 there will be 75 millions of blind people and more than 200 millions visually impaired. However, we should note that not only these people would benefit from a personal visual assistant, but also people with common visual capabilities performing specific tasks (such as firefighters, police or tourists). A personal assistant that helps with the localization and navigation in unknown, difficult or not frequently visited environments, with recognizing a building or with reading a sign in an unknown language would be of great interest.

Different wearable navigation systems have been proposed in the literature for visually impaired people, and a detailed survey is provided in [1]. Currently, there is an increasing interest in devising robotics systems for personal assistance. For instance, [2] presents a robotics proposal for guiding visual-impaired people and for people safety in rescue operations using portable laser-scanner systems on the head. Previous vision-based methods adapted to human navigation and guidance considered stereo vision [3], or wireless communication

This work was supported by project VINEA DPI2012-31781 and FEDER.

technology [4]. Another interesting related proposal is the vision based navigation assistant presented in [5], which builds a topological map using a system composed of four uncalibrated cameras mounted on the user's shoulders. However, the capabilities of these systems are still quite limited and they do not provide enough level of performance as would be required by the possible end users. Additionally, future applications envisioned by society are even more and more demanding in terms of the quality and quantity of the information gathered from the environment by the assistive system.

In this framework, usual approaches consider standard sensors to gather the input information, such as ultrasonic sensors, conventional cameras or compasses. The perception results are also greatly improved by sensor fusion, and relevant advances have been addressed in the last decades in this topic. Nevertheless, we think that long term research should also consider unconventional vision devices to expand the system's potential and to overcome the gap between the research results and the end user demands. Thus, the main goal is to investigate the possibilities of unconventional vision sensors, currently an open research issue, due to the great amount of information they provide and to their low cost and good miniaturization perspectives. A spectacular increase in low-cost computational power has also taken place, making possible the processing of massive sensorial information, opening applications that were inconceivable a few years ago.

The challenge of developing effective systems should be supported by the research of computer vision techniques as part of a personal assistant based on visual localization and scene understanding, which are complementary tasks that can help each other. This will bring new research opportunities but will also raise some challenges. The main difficulties basically consist of the usage of unconventional sensors, offering interesting advantages but also important issues, and the fact that it is a person (or humanoid robot) who wears and moves the sensor, introducing a source of uncertainty because of the unpredictability of human movement, different to the smooth trajectories that mobile robots usually follow.

In this paper, we present a brief survey of some approaches we are working on aiming at expanding the possibilities with the goal of wearable vision systems development for personal assistance. Some of the ideas presented are still in progress and are framed in long term research lines. In particular, we present and discuss three results of our work: (1) Omnidirectional vision for scene layout recovery [6], which is key issue for human navigation and scene understanding. (2) The use of range cameras is studied to improve the detection of obstacle-free paths. (3) The combination of cameras with visible laser for depth capture in flexible configuration.

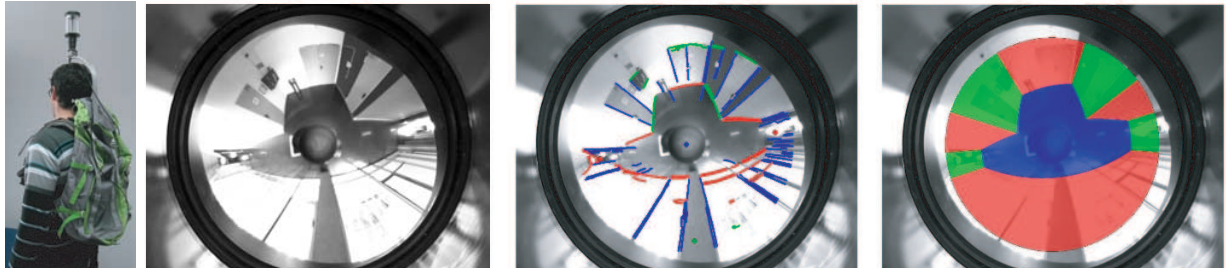

Fig. 1. From left to right: Backpack used for omnidirectional image acquisition. Example of input image fed into our algorithm. Extracted lines and estimated vanishing points classified according to the 3 main directions. Last image shows the floor and vertical walls segmentation obtained with our method.

## II. Scene layout from an omnidirectional view

Conventional cameras are poor imitations of the human eye, which make them restrictive for many applications. Omnidirectional cameras are becoming increasingly popular in computer vision and robotics because their relevant properties. A wide field of view naturally enables a system to *see* in all directions, to detect obstacles or landmarks for navigation, and allows much more robust ego-motion estimation. Moreover, omnidirectional systems capture complete environment information with a reduced number of viewpoints, and therefore a reduced size of the visual memory is enough. But on the other hand, they also raise important geometric and imaging issues which are still open.

Obtaining structural distribution of a scene from an image is an easy task for anyone. However, it is not simple for visually impaired people. Knowing structural limits of the environment is the first step of any autonomous navigation system. Hence the goal is to recover the spatial layout of indoor environments from omnidirectional images assuming a Manhattan world structure. We propose a method for scene structure recovery from a single image. This method is based on line extraction for omnidirectional images, line classification, and vanishing points estimation combined with a hierarchical expansion procedure for detecting floor and wall boundaries.

The works in the literature addressing the problem of spatial layout recovery have been proposed for images acquired by conventional cameras, and most of them work under the Manhattan-world assumption [7]. In general, a conservative spatial layout is enough for defining a navigability map of the environment, and this can be a powerful tool that provides very useful information for performing tasks such as navigation or obstacle detection. Therefore, rather than a precise and detailed map of the scene, we focus on providing a conservative map in which the distribution of the different elements of the scene are classified as floor or walls.

### A. Single view spatial layout recovery

In this section, we briefly describe the proposed algorithm to come up with the spatial layout of the scene from a single omnidirectional image. We start extracting lines from the image, which are then classified according to their orientation in order to carry out the estimation of vanishing points [8]. Combining this information with a set of geometrical constraints we generate hypothesis about the floor contour.

From the classified lines, which are conics in this kind of images, a set of points is selected. These points are used to fit conic lines which represent plausible wall-floor boundaries. Then, a conservative four walls-room hypothesis is generated by selecting the four most voted conic lines. Finally, the initial hypothesis is expanded, according to the image data distribution, to obtain a representative hypothesis. This is carried out by replacing the initial floor contours with a set of appropriate conic lines so that successive hypotheses approximate better the actual shape of the scene. Finally, the output of the proposed algorithm is the spatial layout.

### B. Matching-free sequential hypothesis propagation

This previous single image based algorithm shows good performance and it is robust to occlusions of the scene contours. However, depending on the complexity of the scene, misclassifications can occur. Each single omnidirectional image independently provides a useful hypothesis of the 3D scene structure. In order to enhance robustness and accuracy of this single image-based hypothesis, we extend this estimation with a new matching-free homography-based procedure applied to the various hypotheses obtained along the sequence images.

This approach relies on the homography computation of the floor across the views. It can be demonstrated that considering planar motion and taking into account the estimated vanishing points reduces the minimal set of lines required to compute the homography from $4$ to $1$. Then the exhaustive search in the number of samples, instead of a random search, is feasible in practice, without requiring any prior matching. This homography parametrization allows the design of a matching-free method for spatial layout propagation along a sequence of images. The last step of our method compares hypotheses obtained for the sequence in order to obtain an averaged hypothesis which best fits the set of floor contours.

### C. Experimental results

The proposed method has been tested with a vision system composed by a hypercatadioptric camera attached to a backpack carried by the user (Fig. 1). An example of the results obtained is presented in Fig. 1. It corresponds to a long hall section with an enlargement in one of its sides. Despite the high level of luminosity coming through the window, our method is able to identify every element of the scene.

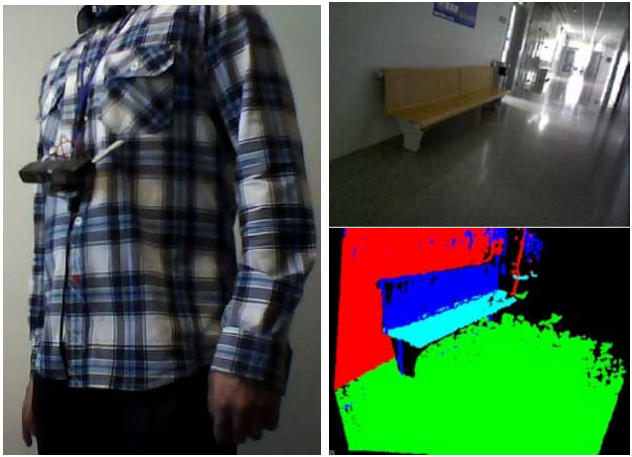

Fig. 2. Example of experimental setup and a resultant planar segmentation.

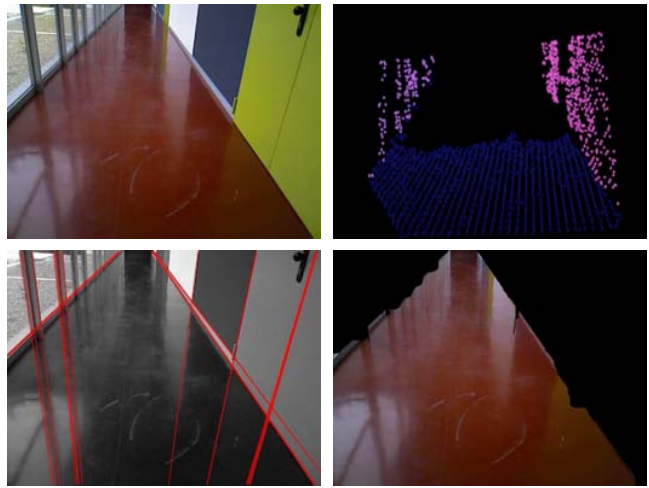

Fig. 3. Experimental results with range camera. Top-left: the original image. Top-right: The point-cloud segmentation. Below-left: Lines extracted. Below-right: The final result after the floor expansion.

## III. Range cameras for navigation aid

Most mobile robots trust on range data for obstacle detection. Popular sensors based on range data are ultrasonic sensors, radar, stereo vision and infrared range sensors. These sensors are able to measure the distance from the sensor to all surrounding obstacles. Obviously, none of these sensors provides perfect results. Ultrasonic sensors have poor angular resolution. Radar works better than ultrasonic sensors but they are more complex and they may have interference problems indoors. Stereo vision needs textured environment and infrared range sensors fail with solar lighting.

In this section, we present a novel method that combines the short-range 3D point-cloud processing and long-range basic image processing for obtaining the free space on the floor. The combination of these processes yields a robust system that works in the presence of severe shadows and reflections, which are common in practice. The sensor chosen is an infrared camera combining monocular vision with range information. This camera provides a point-cloud which contains RGB and 3D information of the scene. The main steps of the proposed algorithm are described next.

### A. Point-cloud analysis

The point-cloud of a scene provided by the range sensor contains a huge amount of 3D information. In order to process the point-cloud as fast as possible, we downsample the point-cloud before the segmentation step. At this point we identify the most representative planes of the scene via a RANSAC procedure. Once we have detected the planes, we classify them by analysing its normal vector and obtain the floor. An example of this classification is shown in Fig. 2. The method works properly indoors and it is robust to lighting changes. On the other hand, it has some limitations: It is susceptible to sunlight and the maximum distance is around 3.5 meters.

### B. Monocular vision analysis

The infrared sensor of the range camera has several limitations as pointed out above and we use the monocular camera to improve the results. In particular, the ground plane detected from the infrared sensor is used for obtaining the whole plane of the ground in the RGB image. A significant amount of research has focused upon image segmentation problem [9] [10]. Here, we combine the depth information with the image using RGB and HSI color spaces and geometry image features. First, we perform a seeded region growing algorithm where the seed belongs to the 3D floor's plane of the depth point-cloud. In order to reduce the shadows and reflections influence, we apply a mean-shift filter to a pyramid image.

The next step is to compare the lighting and hue channel of the homogenized image with the seed. The floor is not homogeneous so the seed will have a variety of hue values. So, we compare each part of the image satisfying the first criterion, with each hue value of the seed. We carry this task out evaluating how well the pixels fit a histogram model. Pixels which satisfy this criterion will become seeds. Finally, we propose a polygon-based region growing step. We use the Probabilistic Hough Line Transform in order to extract lines and extend them till the image's border. Once we have the image segmented by polygons and the seeds distributed along the putative floor, the last step is to grow those seeds. Each polygon which has one or more seeds inside, will be labelled as floor.

### C. Experimental evaluation

The hardware used is an Asus Xtion Pro live camera that hangs from the user's neck as shown in Fig. 2. The range camera will be slightly tilted towards the ground in order to detect the closest obstacles. To evaluate the performance of our algorithm, we have tested it in different kind of corridors exhibiting a wide variety of different visual characteristics and lighting conditions. Fig. 3 presents an example of a typical corridor image, notice the floor reflectivity and lighting conditions.

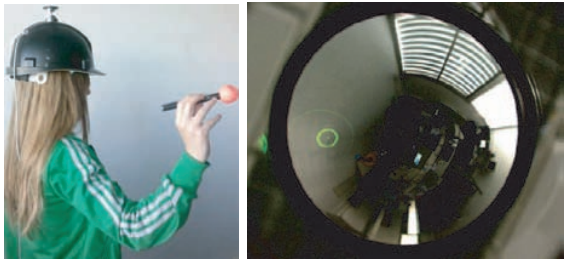
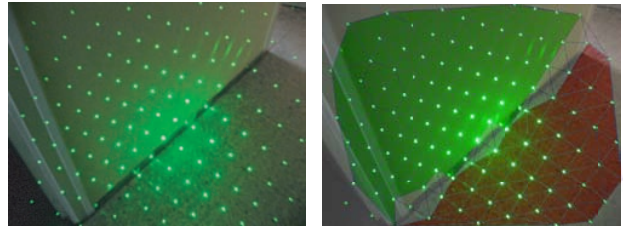

Fig. 4. Wearable vision system with laser in hand (left). Omnidirectional view with the projected conic pattern light for image processing (right).

Fig. 5. Image with dotted pattern laser projected indoors (left). Resultant floor and wall homography-based segmentation (right).

## IV. VISION-BASED PERCEPTION WITH VISIBLE LASER

Structured light is an active method that provides 3D information from images of the scene by projecting synthetic features with a low-cost light emitter. In the previous section, infrared system was considered, in order to develop a more flexible system, we study the use of visible laser in the context of wearable personal assistance systems.

### A. Conic laser pattern

Traditionally, structured light methods consider a rigid configuration, where the position and orientation of the light emitter with respect to the camera are known beforehand. We have developed a new omnidirectional flexible structured light system which overcomes the rigidness of the traditional structured light systems. We propose the use of an omnidirectional camera combined with a conic pattern light emitter in hand (Fig. 4). Since the light emitter is visible in the omnidirectional image, the computation of its location is possible. With this information and the projected conic in the omnidirectional image, we are able to compute the 3D conic reconstruction.

Our approach [11] combines the omnidirectional image and a virtual image generated from the light emitter. With our method, we obtain the depth and orientation of the scene surface where the conic pattern is projected. The long term application of this structured light system in flexible configuration is a wearable omnicamera with a laser in hand for visual impaired personal guidance.

### B. Dotted laser pattern

Alternatively, we study the use of dotted laser patterns avoiding the need of system calibration and 3D reconstruction. The setup consists of a green laser in hand, projecting light points following a squared pattern, and a camera mounted on the person. We have designed a homography-based approach that uses the image of the dotted pattern projected in the scene to obtain plane segmentation. This is an efficient procedure that provides fast and robust obstacles detection without constraining to a fixed configuration between camera-laser. Fig. 5 shows an example of the results obtained. The method works correctly with changes in illumination conditions and shadows. Therefore, we consider this setup as an interesting choice in the framework of personal guidance systems.

## V. DISCUSSION AND CONCLUSION

Several unconventional vision-based systems have been studied for the particular task of scene analysis for human navigation guidance. Omnidirectional views provide a wide field of view and we presented a method for spatial layout recovery from a single omnidirectional image based on conic lines classification and its propagation along a sequence of images skipping the prone to error and computational expensive matching algorithms. Range infrared cameras are a powerful tool providing 3D information at close range, robust to lighting changes, glows and reflections. We combine this information with vision to extend the obtained information along all the visible scene. Finally, we study the use of visible laser in the framework of obstacle detection.

These are just some examples of the possibilities of wearable vision systems for personal assistance. These systems can be easily carried by a person and are able to guide a visually impaired people by, for instance, audio instructions. However, none of them will guarantee total absence of failures, and many issues will emerge, opening novel research lines in order to address this challenge.

## REFERENCES

[1] D. Dakopoulos and N. Bourbakis, "Wearable obstacle avoidance electronic travel aids for blind: A survey," *IEEE Trans. on Systems, Man, and Cybernetics, Part C*, vol. 40, no. 1, pp. 25–35, 2010.
[2] M. Baglietto, A. Sgorbissa, D. Verda, and R. Zaccaria, "Human navigation and mapping with a 6 DOF IMU and a laser scanner," *Robotics and Autonomous Systems*, vol. 59, no. 12, pp. 1060 – 1069, 2011.
[3] J. M. Saez Martinez and F. Escolano Ruiz, "Stereo-based Aerial Obstacle Detection for the Visually Impaired," in *Workshop on Computer Vision Applications for the Visually Impaired, in ECCV*, Marseille, 2008.
[4] R. Öktem, E. Aydin, and N. Cagiltay, "An indoor navigation aid designed for visually impaired people," *IECON 2008*, pp. 2982–2987, 2008.
[5] O. Koch and S. Teller, "Body-relative navigation guidance using uncalibrated cameras," in *ICCV*, 2009, pp. 1242–1249.
[6] J. Omedes, G. López-Nicolás, and J. J. Guerrero, "Omnidirectional vision for indoor spatial layout recovery," in *Frontiers of Intelligent Autonomous Systems*. Springer, 2013, pp. 95–104.
[7] D. Lee, M. Hebert, and T. Kanade, "Geometric reasoning for single image structure recovery," in *IEEE Conference on Computer Vision and Pattern Recognition*, June 2009.
[8] J. Bermudez-Cameo, L. Puig, and J. J. Guerrero, "Hypercatadioptric line images for 3D orientation and image rectification," *Robotics and Autonomous Systems*, vol. 60, pp. 755–768, 2012.
[9] Y. Li and S. Birchfield, "Image-based segmentation of indoor corridor floors for a mobile robot," *IEEE/RSJ International Conference on Intelligent Robots and Systems*, pp. 837–843, 2010.
[10] H. Dahlkamp, A. Kaehler, D. Stavens, S. Thrun, and G. Bradski, "Self-supervised monocular road detection in desert terrain," *Proceedings of Robotics: Science and Systems*, August 2006.
[11] C. Paniagua, L. Puig, and J. J. Guerrero, "Omnidirectional structured light in flexible configuration," *Technical Report, I3A*, 2013.

# Learning the Relation of Motion Control and Gestures Through Self-Exploration

Saša Bodiroža*, Aleksandar Jevtić†, Bruno Lara‡, Verena V. Hafner*

*Institut für Informatik, Humboldt-Universität zu Berlin
Berlin, Germany
Email: {bodiroza,hafner}@informatik.hu-berlin.de

†Robosoft
Bidart, France
Email: aleksandar.jevtic@robosoft.com

‡Cognitive Robotics Group, Faculty of Science
Universidad Autonoma del Estado de Morelos
Cuernavaca, Mexico
Email: bruno.lara@uaem.mx

*Abstract*—This work presents an action execution system that uses as foundation very basic sensorimotor schemes. These schemes are learned as a product of the interaction with an agent with its environment. In the example experiment, a mobile agent learns an association between changes in its sensory perception and the movements it performs. Once it has this acquired knowledge, the agent is then capable of performing a mirror action to match an observed gesture. This is seen as a first step toward learning motor control strategies for a robot control task.

## I. Introduction

Human-robot interaction is usually confined within the research laboratories. As a step of moving robots outside of laboratories and into the real world, a robot waiter scenario is proposed, which is based on integration of multiple research areas into a final platform [8]. Toward this goal, robot control based on gesture recognition was considered [3, 5]. This paper presents the recognition and execution of direction gestures, namely left, right, up and down, as a mean for basic robot control and attention manipulation. The motor control strategy is a result of learning internal models through sensorimotor exploration, as demonstrated by Dearden and Demiris [4]. Changes in sensory input, represented by the changes of the location of the person's right hand, are observed during learning solely as a result of the robot's exploration.

The goal of the presented experiment is to display the viability of learning motor control based on self-exploration. The presented approach could also be applied to other scenarios, such as learning tool-use [11], learning how to focus eyes on a particular object, learning relations between a motor command and position of the end effector in space – in general the result of particular motor actions on the robot's environment. A relation between a motor command and its consequence would be learned from the collected data. This relation can be also used in the other direction – if the robot has a particular goal or state that needs to be reached, the relation could predict what is the appropriate motor command that would take the robot to the desired state.

The topic of learning inverse kinematics has been an area of interest for some time. D'Souza et al. presented a method for learning inverse kinematics for a robot arm using locally weighted projection regression [6]. A similar approach is presented by Lapreste et al. [9]. However, the approach presented here does not employ machine optimization strategies, yet achieves sufficient results for the rotation task. Previous work by Schillaci et al. showed that a similar approach can be used to learn a motor control strategy for pointing or reaching for a robot arm using motor babbling during learning and internal models for prediction [11]. An argument for importance of internal models in motor learning and possible approaches is given by Wolpert et al. [15].

## II. Internal Models and Gesture Recognition

### A. Internal Models

Internal models represent a theoretical concept, consisting of a pair of inverse and forward models, represented in figure 1. Depending on a problem, an inverse model can predict a motor command $M_t$ that leads the system from the current state $S_t$ to a desired state $S_{t+1}$, or, based on the observed change of states from $S_t$ to $S_{t+1}$, predicts an appropriate motor command $M_t$. The forward model performs an internal simulation and predicts the state $S_{t+1}^*$ that would be the result of the motor command $M_t$ in the state $S_t$.

Similar to our approach, Dearden and Demiris perform learning of forward models for action execution [4]. In their work, a mobile robot observes the motion of its gripper while sending to it random motor commands. A forward model is obtained, that establishes the connection between motor commands and the changes in the visual space caused by those motor commands. This way, the system is able to imitate human movements.

Akgün et al. [1] show how an action generation mechanism can be used for action recognition. They developed an online recognition system, that was able to recognize a reaching action before it was fully executed.

Haruno et al. [7] and Wolpert and Kawato [16] present evidence for development of multiple, tightly-coupled inverse and forward models. The forward model predicts the result of the motor command generated by the inverse model. The selection of the best inverse-forward model pair is done through comparison of predictions of all forward models to the expected result. Schillaci et al. [12] used multiple internal models to perform recognition of human behavior, where each internal model encodes an action. Blakemore et al. [2] present how a difference in the prediction of the forward model and the perceived sensory input helps a person discriminate a self-induced sensation (e.g. self-movement of the eye) from a sensation induced by others (e.g. moving the eye by pressing on the eyelid). Furthermore, Takemura and Inui [14] present a model for the development of internal models for reaching movement, inspired by infant development.

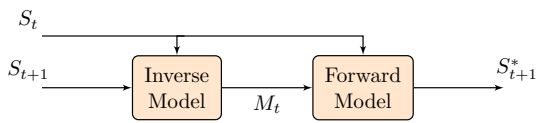

Fig. 1. Internal models. An inverse model predicts a motor command $M_t$ that leads the system from the current state $S_t$ to $S_{t+1}$. A forward model predicts the state $S_{t+1}^*$ based on the current state $S_t$ and the motor command $M_t$.

### B. Gestures

As presented in [13], certain aspects of robot behavior have been identified in order to increase intuitiveness of interaction, such as enhanced feedback from the robot. Robot pointing was used to manipulate human attention. On the other side, gestures can be used to manipulate robot attention and to indicate intentions of a human participant. Therefore, robust gesture recognition is important for a satisfying, real-life human-robot interaction.

Gesture recognition is based on detection of the position (static) or the motion (dynamic) of human body parts, usually arms, hands and legs. It has numerous applications, including sign-language for hearing-impaired people, computer interfaces, natural and intuitive human-robot interaction, gaming industry, system remote control, among others. Various tools have been used for gesture recognition, based on the approaches ranging from statistical modeling, computer vision and pattern recognition, image processing, connectionist systems, etc. [10].

However, learning from self-exploration to observe and reproduce certain actions from gestures of others has not been done before, apart from work of Dearden and Demiris [4]. The proposed model learns to execute actions associated with directional gestures. The learned motor controls are a result of a motor babbling process, during which a random motor command is generated and a change in the sensory input is observed and stored.

### III. Proposed model

The proposed model uses as sensory situation the detected coordinates of the arm of a person as tracked by a Kinect. The motor commands are fixed movements of the mobile platform.

In our model, we want to perform the fusion of sensorimotor information. To achieve this, the system needs to collect, for each time step, a vector of the form:

$$(x, y, z); M \qquad (1)$$

where $(x, y, z)$ represent the coordinates of the hand detected by the Kinect and $M$ represents a random motor movement. This movement can be either in the left-right plane, performed by the robot as a rotation for a random angle, within $[-27°, 27°]$ or in the up-down plane, performed as a tilt angle in the Kinect within $[0°, 16°]$ (ranges were selected to always have the person's upper body visible).

Once a database of these associations is collected, it can be used as either a forward or an inverse model, depending on the question asked.

The inverse model predicts a motor command $M_t$, when presented with a change in the sensory situation from $S_t$ to $S_{t+1}$. The forward model, given the current sensory situation $S_t$ and a motor command $M_t$ predicts the new sensory situation after the execution of the command, $S_{t+1}^*$. In the proposed application, that is a gesture-controlled robot, during the learning process random motor commands induce changes in sensory situations, that is in the position of the hand. The model associates the performed motor command $M_t$ with the sensory situation before the execution $S_t$ and after the execution $S_{t+1}$. During execution the model performs a search for a motor command $M_t$ that matches the sensory change, only this time induced by a person by moving their hand.

It could be said, that the model learns how to execute actions. It learns how the self-motion corresponds to changes in the world, which is then applied during the execution, when it is required to reproduce an action that resulted in the observed change. The resulting behaviour can be also seen as an attention manipulation system, where the robot turns following the motion of a hand.

### IV. Experiments and Results

Two experiments were performed to learn the mapping between the motor commands and changes in the sensory situation. A robot platform robuLAB, displayed in figure 2, was used in the experiments. In the first experiment, the association of "up" and "down" gestures and Kinect's up and down movements was formed, while in the second experiment the association of "left" and "right" gestures and the rotation of the robuLAB platform was learned. The following description shows the outline of both experiments, and "the platform" represents either the Kinect or the robot platform. In the former case, the motor commands were tilting the Kinect, while in the latter they were rotation of the robot platform around the z-axis.

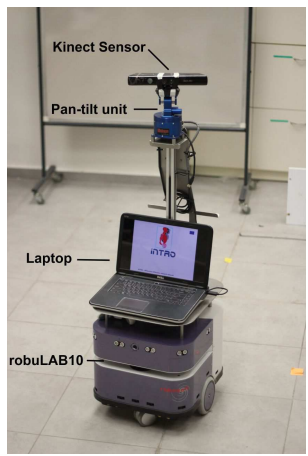

Fig. 2. Experimental platform

A person stood in front of the platform at the beginning of the experiment. The Kinect was tracking the location of the person's right hand, provided by the Microsoft's Kinect SDK. During the learning stage the platform was performing motor babbling, which was a generation of random motor commands. Every motor command induced a change in the sensory situation, that is the change of the 3D location of the person's hand. This change, represented with the $(x, y, z)$ vector of the hand movement in Kinect's frame of reference, corresponds to the rotation angle of the platform. In other words, if the platform rotates to the left, the hand will be seen as moving right. However, during the execution phase, if the platform perceives the hand moving right, it should rotate to the right, instead of left. During the learning process, initial rotation of the platform and the generated random motor command, represented as a rotation angle, was stored, as well as the 3D position of the hand before and after the rotation.

A k-nearest neighbors search algorithm was used for implementation of the inverse model. Theoretically, the initial state $S_t$ represents the initial position of the hand, and the next state $S_{t+1}$ the new position of the hand, resulting from the person's movement. The predicted motor command $M_t$ represents the rotation angle of the robot that it needs to perform in order to compensate for the motion of the person's hand. However, in order to make the implementation more robust with regards to the person's location in space, $S_t$ and $S_{t+1}$ are combined and represented as the difference vector of the new and the initial location of the person's hand.

When the platform observes the movement of the person's arm, it uses the perceived change in the hand's location to predict a motor command that will compensate for the motion of the arm. The resulting action can be seen as recognition of a dynamic gesture and its corresponding motion.

Testing of the algorithm was done using the rotation of the robuLAB platform. Training was performed with 60 points from motor babbling. Testing was performed with hand movements to the left or to the right for 25 times. On average, absolute hand displacement of the user was $(x, y, z) =$

$(0.43, 0.24, 0.06)m, s.d. = (0.13, 0.12, 0.05)$ and the error of the prediction resulted in the absolute mismatch between the initial hand position and the hand position after the rotation of $(x, y, z) = (0.08, 0.24, 0.06)m, s.d. = (0.06, 0.12, 0.04)$. The results show that this approach can be used for learning motor control for rotation based on the depth data of the user's hand.

## V. Conclusion and Future Work

This work presents learning and execution of an inverse-forward model pair to execute actions associated with directional gestures. While only the inverse model was trained, the forward model could be easily added and used to predict the hand location after the robot's movement. This information can be used for error measurement of the prediction and refinement of the initial motion, if the error is higher then a certain threshold. Future work could improve the proposed model to understand pointing gestures, with the goal of learning a control strategy for moving to a specific location. This location would be indicated by a person pointing.

## Acknowledgments

## References

[1] B. Akgün, D. Tunaoğlu, and E. Şahin. Action recognition through an action generation mechanism. In *International Conference on Epigenetic Robotics (EPIROB)*, 2010.

[2] S. J. Blakemore, D. Wolpert, and C. Frith. Why can't you tickle yourself? *Neuroreport*, 11:11–16, 2000.

[3] Saša Bodiroža, Guillaume Doisy, and Verena Vanessa Hafner. Position-invariant, real-time gesture recognition based on dynamic time warping. In *Proceedings of the 8th ACM/IEEE international conference on Human-robot interaction*, HRI '13, pages 87–88, Piscataway, NJ, USA, 2013. IEEE Press. ISBN 978-1-4673-3055-8.

[4] Anthony Dearden and Yiannis Demiris. Learning forward models for robots. In *Proceedings of the 19th international joint conference on Artificial intelligence*, IJCAI'05, pages 1440–1445, 2005.

[5] Guillaume Doisy, Aleksandar Jevtić, and Saša Bodiroža. Spatially unconstrained, gesture-based human-robot interaction. In *Proceedings of the 8th ACM/IEEE international conference on Human-robot interaction*, HRI '13, pages 117–118, Piscataway, NJ, USA, 2013. IEEE Press. ISBN 978-1-4673-3055-8.

[6] A. D'Souza, S. Vijayakumar, and S. Schaal. Learning inverse kinematics. In *Proceedings of IEEE/RSJ International Conference on Intelligent Robots and Systems*, volume 1, pages 298–303, 2001. doi: 10.1109/IROS.2001.973374.

[7] Masahiko Haruno, Daniel M. Wolpert, and Mitsuo Kawato. In David Cohn Michael Kearns, Sara Solla, editor, *Advances in Neural Information Processing Systems*, volume 11, pages 31–37, Cambridge, MA.

[8] Aleksandar Jevtic, Eric Lucet, Alex Kozlov, and Jeremi Gancet. INTRO: A multidisciplinary approach to intel-

ligent Human-Robot Interaction. In *World Automation Congress (WAC), 2012*, pages 1–6, 2012.

[9] J.-T. Lapreste, F. Jurie, M. Dhome, and F. Chaumette. An efficient method to compute the inverse Jacobian matrix in visual servoing. In *Proceedings of IEEE International Conference on Robotics and Automation*, volume 1, pages 727–732, 2004. doi: 10.1109/ROBOT. 2004.1307235.

[10] Sushmita Mitra and Tinku Acharya. Gesture recognition: A survey. *IEEE Transactions on Systems, Man, and Cybernetics, Part C: Applications and Reviews*, 37(3): 311–324, 2007.

[11] Guido Schillaci, Verena Vanessa Hafner, and Bruno Lara. Coupled inverse-forward models for action execution leading to tool-use in a humanoid robot. In *Proceedings of the seventh annual ACM/IEEE international conference on Human-Robot Interaction*, pages 231–232. ACM, 2012.

[12] Guido Schillaci, Bruno Lara, and VerenaV. Hafner. Internal Simulations for Behaviour Selection and Recognition. In AlbertAli Salah, Javier Ruiz-del Solar, etin Merili, and Pierre-Yves Oudeyer, editors, *Human Behavior Understanding*, volume 7559 of *Lecture Notes in Computer Science*, pages 148–160. Springer Berlin Heidelberg, 2012. ISBN 978-3-642-34013-0. doi: 10.1007/978-3-642-34014-7_13.

[13] Guido Schillaci, Saša Bodiroža, and Verena Vanessa Hafner. Evaluating the Effect of Saliency Detection and Attention Manipulation in Human-Robot Interaction. *International Journal of Social Robotics*, 5(1): 139–152, 2013. ISSN 1875-4791. doi: 10.1007/ s12369-012-0174-7.

[14] Naohiro Takemura and Toshio Inui. A Developmental Model of Infant Reaching Movement: Acquisition of Internal Visuomotor Transformations. In Rubin Wang and Fanji Gu, editors, *Advances in Cognitive Neurodynamics (II)*, pages 135–138. Springer Netherlands, 2011. ISBN 978-90-481-9694-4. doi: 10.1007/978-90-481-9695-1_ 21.

[15] Daniel M. M. Wolpert, Zoubin Ghahramani, and J. Randall Flanagan. Perspectives and problems in motor learning. *Trends in cognitive sciences*, 5(11):487–494, Nov. 2001. ISSN 1879-307X.

[16] D.M. Wolpert and M. Kawato. Multiple paired forward and inverse models for motor control. *Neural Networks*, 11(7-8):1317–1329, 1998. ISSN 0893-6080. doi: 10. 1016/S0893-6080(98)00066-5.